\documentclass[a4paper,fleqn]{cas-dc}

\usepackage[numbers]{natbib}
\usepackage{xcolor}
\usepackage{colortbl}
\usepackage{longtable}
\usepackage{xcolor}
\usepackage{array}
\usepackage{booktabs}
\usepackage{lipsum}
\usepackage{blindtext}
\hypersetup{
    colorlinks=true,
    linkcolor=blue,
    filecolor=magenta,      
    urlcolor=cyan,
    pdftitle={Overleaf Example},
    pdfpagemode=FullScreen,
    }

\usepackage{fancyhdr}
\begin{document}
\let\WriteBookmarks\relax
\def\floatpagepagefraction{1}
\def\textpagefraction{.001}


\shortauthors{M Gantiva et~al.}

\title [mode = title]{Coordinated control of multiple autonomous surface vehicles: challenges and advances - a systematic review}                      
\tnotemark[1]

\tnotetext[1]{This work was supported by the Agencia Estatal de Investigación (AEI) through the Project AQUATRONIC under Grant PID2021 126921OA-C22, and through the Project ECOPORT under Grant TED2021-131326A-C22.}


%
\author[1]{Manuel Gantiva Osorio}

\cormark[1]


\ead{megantiva@uloyola.es}



\affiliation[1]{organization={Department of Engineering, Universidad Loyola Andalucía},
    city={Seville},
    postcode={41704}, 
    country={Spain;}}

\author[1]{Carmelina Ierardi}
\ead{cierardi@uloyola.es}



\author[1]{Isabel Jurado Flores}
\ead{ijurado@uloyola.es}

\author[1]{Mario Pereira Martín}
\ead{mpereira@uloyola.es}

\author[1]{Pablo Millán Gata}
\ead{pmillan@uloyola.es}

\cortext[cor1]{Corresponding author}



\begin{abstract}
The increasing use and implementation of Autonomous Surface Vessels (ASVs) for various activities in maritime environments is expected to drive a rise in developments and research on their control. Particularly, the coordination of multiple ASVs presents novel challenges and opportunities, requiring interdisciplinary research efforts at the intersection of robotics, control theory, communication systems, and marine sciences. The wide variety of missions or objectives for which these vessels can be collectively used allows for the application and combination of different control techniques. This includes the exploration of machine learning to consider aspects previously deemed infeasible. This review provides a comprehensive exploration of coordinated ASV control while addressing critical gaps left by previous reviews. Unlike previous works, we adopt a systematic approach to ensure integrity and minimize bias in article selection. We delve into the complex world of sub-actuated ASVs with a focus on customized control strategies and the integration of machine learning techniques for increased autonomy. By synthesizing recent advances and identifying emerging trends, we offer insights that drive this field forward, providing both a comprehensive overview of state-of-the-art techniques and guidance for future research efforts.

\end{abstract}




\maketitle

\section{Introduction} \label{sec:introduction}
In recent years, the deployment of ASVs has gained considerable momentum across various fields, including marine exploration, environmental monitoring, maritime security, and offshore operations. These unmanned platforms offer the potential to revolutionize our approach to tasks that were once reliant on crewed vessels, enabling greater efficiency, reduced operational costs, and increased safety in challenging maritime environments. As the capabilities of individual ASVs continue to evolve, the focus has shifted towards harnessing their collective potential through coordinated control strategies. The coordinated operation of multiple ASVs presents a unique set of challenges and opportunities that necessitate interdisciplinary research efforts at the intersection of robotics, control theory, communication systems, and marine science  \citep{Jin2016}, \citep{Li2018Finite}, \citep{Peng2013AdaptiveDynamic}, \citep{Shojaei2015}.

The concept of coordinating multiple ASVs goes beyond the mere aggregation of individual vehicles; it involves orchestrating their actions to achieve common objectives while adapting to dynamic environmental conditions and mission requirements. This requires the development of advanced algorithms that enable ASVs to collaborate, communicate, and adapt in real-time, fostering a synergy that surpasses the capabilities of isolated units. Coordinated control of ASVs not only enhances the efficiency and effectiveness of various marine applications but also opens up new avenues for scientific exploration and data collection in remote or hazardous regions of the world's oceans.

In this paper we present a comprehensive review that delves into the intricate landscape of coordinated control for systems of multiple ASVs, proposing a systematic approach.
While acknowledging the contributions of traditional prior reviews (\cite{gu2022overview}, \cite{karimi2021survey}, \cite{peng2020overview}, \cite{qiao2023survey}, \cite{thombre2020review}, \cite{zhang2021survey}), as shown in Table \ref{fig: previous reviews}, this systematic review aims to push the envelope by introducing fresh perspectives that tackle the challenges associated with coordinated ASV control, ranging from communication constraints and collision avoidance to decentralized decision-making and adaptive mission planning. Furthermore, we highlight recent advances and innovations in control strategies and sensing technologies that are propelling the field forward. By surveying the state-of-the-art methods and discussing emerging trends, we aim to provide a comprehensive overview of the current research landscape and inspire further advancements in the field of coordinated ASV control.

The specific contributions of this review are as follows:
\begin{itemize}
    \item It presents a systematic review of the coordinated control of multiple underactuated ASVs, filling a notable gap in the literature where traditional reviews exist but lack this specific, systematic approach.
    \item The review incorporates, among others, studies that explore the use of machine learning techniques in ASV control coordination.
    \item A broad range of characteristics from each selected article has been extracted and compiled into tables to simplify the reading and information search process, making it as straightforward and visual as possible.
    \item The review has been updated to include information up until September 2023.
\end{itemize}

\begin{table}[htbp]
\caption{Summary of contributions in major reviews on coordinated ASVs control}
	\centering
            \scriptsize{
		\begin{tabular}{{>{\centering\arraybackslash}m{1.165em}
                >{\centering\arraybackslash}m{3.15em}
                >{\raggedright\arraybackslash}m{21.50em}}}\toprule
		   \textbf{Ref}& \textbf{Year} & \textbf{Focus-Main contribution} \\ \midrule
            \cite{peng2020overview} & 2020 & Overview of   recent advances in coordinated control of multiple ASVs, addressing challenges in motion control and fleet   operations. Discusses coordinated control methods and recent results,   highlighting future research directions for enhanced autonomy and efficacy. \\

             &  &  \\
            
            \cite{zhang2021survey} & 2021 & Comprehensive   overview of R\&D advances in Maritime Autonomous Surface Ships (MASS) industry and collision-avoidance   navigation. Analysis of brain-inspired cognitive navigation, e-navigation   technologies, highlighting trends in maritime collision-avoidance systems. \\

              &  &  \\
            
             \cite{karimi2021survey} & 2021 & Review   highlights recent offshore mechatronics advancements, focusing on guidance   and control methodologies for marine robotic vehicles. Covers fuzzy-based,   neural network-based, dynamic surface, feedback, and sliding model control   methods for maneuvering, path following, and formation control. \\

               &  &  \\
 
            \cite{thombre2020review} & 2022 & Overview   of perception system requirements for autonomous ships, emphasizing sensor   fusion with AI techniques. Integration of Global Navigation Satellite System (GNSS), Inertial Measurement Unit (IMU), visual, audio, and   remote-sensing sensors discussed. AI methods like deep learning identified   for tasks like abnormality detection and vessel classification. \\

              &  &  \\
 
            \cite{gu2022overview} & 2023 & Survey of recent advances in Line-Of-Sight (LOS) guidance for Autonomous Marine   Vehicles (AMV), crucial for motion control. Covers LOS guidance laws for   path following and coordinated path following of multiple AMVs, highlighting   future research directions. \\

              &  &  \\
 
            \cite{qiao2023survey} & 2023 & Survey   outlines advancements in maritime autonomous systems, particularly in sensor   technology and deep learning for ASVs. It discusses current challenges and   provides insights into future research directions \\

              &  &  \\

            own & 2024 & Our   paper focuses on underactuated surface vessels and investigates coordinated   control techniques specifically tailored to their unique characteristics. We   explore methods, their pros and cons, applications and detail the coordinated   control approaches utilized in each scenario. Additionally, we examine which   techniques incorporate machine learning applications for enhanced autonomy.\\

         \bottomrule
	\end{tabular}}
\label{fig: previous reviews}
\end{table}

Given Table \ref{fig: previous reviews}, the factors that differentiate this review from \citep{peng2020overview} and \citep{qiao2023survey}, which are the ones that present the most points in common with the present work, are presented below.

Regarding \citep{peng2020overview}, the factors that differentiate it from the present review, in which a novel contribution is presented, are:
\begin{itemize}
     \item In \citep{peng2020overview} the systematic review approach is not used, so the selected papers may suffer from arbitrariness, bias, or the selection may not be complete.
     \item The review \citep{peng2020overview} was published in 2020, and considering Figure \ref{FIG:percentage_controls}, most of the articles that address the coordinated control of ASVs were published in the following years.
     \item Studies dealing with the use of machine learning techniques in the control coordination of ASVs are not included in \citep{peng2020overview}, unlike in the present work.
     \item In \citep{peng2020overview}, fully actuated ASVs are also consider, not only underactuated ones. Consequently, it delves into this particular subject to a lesser extent than our work.
     \item In \citep{peng2020overview}, each paper included in the review is not analyzed in depth, unlike in the present study, in which a lot of information about communication, disturbances, simulations/experiments, among others, is extracted.
     \item Almost no experimental results are analyzed in \citep{peng2020overview}. On the other other side, this work includes a detailed analysis of the experimental validation and applications (Table \ref{fig: data_experimental} and Table \ref{fig: data_model_experimental}).
  \end{itemize}    

Regarding now the differential factors with \citep{qiao2023survey}, the present study provides the following novelties:
 \begin{itemize}
     \item In this instance as well, in \citep{qiao2023survey}, the systematic review approach is not used, so the selected papers may suffer from arbitrariness, bias, or the selection may not be complete.
     \item In this context as well, the control of fleets with underactuated ASVs, although it is included in the review \citep{qiao2023survey}, is not its main focus. Note that the survey \citep{qiao2023survey} covers topics such as navigation systems that include perception with computer vision or radars, path generation, estimation and control of an ASV and, on top of that, estimation and control of ASVs fleets.
     \item In \citep{qiao2023survey}, only articles in which deep learning techniques are applied are considered, while in the present work this is just a feature of the control system that may or may not be present.
     \item The main objective in \citep{qiao2023survey} is to identify the gap between research in deep learning techniques and its applicability to ASVs, while this work presents the latest advancements in the coordinated control of multiple ASVs, among which may be deep learning techniques and their applications.
 \end{itemize}

The rest of this systematic review is organized as follows: Section \ref{sec:review_method} provides the guidelines on how the systematic review was conducted. Section \ref{sec:report} presents the answers through discussions for each of the posed questions. In Section \ref{sec:gap_future_research}, the identified gaps and future works are presented. Finally, Section \ref{sec:conclusions} offers some conclusions.

\section{Review method} \label{sec:review_method}
In this section, we outline the methodological framework employed to conduct our systematic review focused on the coordinated control methods for multiple ASVs. Given the complexity and breadth of the topic, it has become imperative to establish a rigorous and structured approach to identify, select, and analyze relevant literature. To this end, our systematic review has been developed in accordance with the guidelines set forth by \cite{brereton2007lessons} and \cite{ierardi2019distributed}, ensuring a methodologically sound and comprehensive examination of the field.

After reviewing more than 2300 potential articles, merely 125 primary studies were determined to meet the established criteria. This data is subsequently compiled into different feature tables, serving as an effective method to achieve a comprehensive, thorough, and visual understanding of the selected subject. Upon concluding the process, the initially posed research questions are addressed, offering a complete insight into the topic. 

The systematic review is structured into three sequential stages, each further divided into specific sub-stages, as outlined below:
\begin{enumerate}
    \item Planning the review
    \begin{itemize}
        \item identification of the need
\item research questions
\item review protocol
\item  evaluating protocol
    \end{itemize}
    \item Conducting the review
    \begin{itemize}
        \item selection of primary studies
\item study quality assessment
\item extraction and synthesis of data
    \end{itemize}
    \item Reporting the review
    \begin{itemize}
        \item specifying dissemination mechanisms
\item formatting the main report
\item evaluating the report
    \end{itemize}
\end{enumerate}

\subsection{Planning the review} \label{subsec:planning}
The planning phase marks the initial step of the systematic review, laying the groundwork for the entire process. At this juncture, essential tools are established, including the boolean function, criteria for inclusion and exclusion, selection of databases for the research, and crucially, the creation and appraisal of a protocol overseeing all stages.

\textbf{- Identification of the need:} 
The motivation for embarking on a systematic review primarily emerges from the extensive nature of the research topic, necessitating a precise and rigorous approach for the accurate extraction of information. In the field of engineering, adopting such a methodology is uncommon due to its scientific rigor and the complexity involved in its execution. 

Presently, there is a lack of systematic reviews concerning coordinated control techniques for ASV. Moreover, the traditional reviews that do exist differ from the approach proposed in this article, with all differences detailed in Table \ref{fig: previous reviews}.

\textbf{- Research questions:}
After pinpointing the specific topic, certain criteria assist in the precise formulation of research questions. The PICOC (Population, Intervention, Comparison, Outcome, Context) criteria, widely utilized across various fields, have been considered in this study. We have taken into account the criteria outlined in \cite{ierardi2019distributed} and customized them to suit our specific situation.
In this instance, only a selection of these criteria were employed to craft and address the questions this systematic review aims to resolve. Section \ref{sec:report} features a discussion where the corresponding answers are scrutinized and deliberated. The questions devised for this study are enumerated below:\\
RQ.1: What coordinated control techniques are used in multiples autonomous surface vehicles?\\
RQ.1.1: What are the limitations and advantages of the different techniques?\\
RQ.2:Which are the most important characteristics about techniques that have been experimentally tested?\\
RQ.3:Which coordinated control methods are used in each scenario, taking into account the influence of method configuration and communication topology?\\
RQ.4: What machine learning techniques, including deep learning and other neural networks, are used and for what purposes?

\textbf{- Review protocol:}
The review protocol comprises a series of guidelines and standards that must be adhered to throughout all phases of the review, aiming to minimize bias and enhance the objectivity of the systematic review. In biosciences, such protocols are often documented in prospective registries like PROSPERO (https://www.crd.york.ac.uk/ prospero/). Regrettably, equivalent registries are not available in the field of automation and control, so the following stages are mainly based on the protocol presented in \cite{ierardi2018guidelines}.

For the systematic review, the clear presentation and development of the protocol are crucial, given that at least two individuals are involved in drafting the review. A prevalent approach, albeit time-intensive, involves two independent reviewers conducting and reporting the review phases separately, followed by a comparison and discussion of the results they each obtain. The primary aim of this exhaustive work is objectivity.

A fundamental part of the protocol involves devising a boolean function that encompasses all terms pertinent to the selected theme, encapsulating synonyms and related terms to the key words of interest. Conducting this keyword-based research necessitates an in-depth exploration of the subject to identify the terms most commonly employed by authors.

The boolean function formulated for this study is as follows:\\
(ASV OR "surface vehicle" OR USV OR UMV OR "marine vehicle" OR "surface vessel" OR "surface craft" OR "marine craft" OR "Surface ship" OR "underactuated ship" OR "under-actuated ship" OR "marine vessel" OR "marine offshore" OR "Watercraft" OR "Water-craft" OR "sailcraft" OR "boat") \textbf{AND}\\
("flocking control" OR "Swarm control" OR "coordinated control" OR "formation control" OR "Cooperative control" OR "formation tracking" OR "control for target tracking" OR "tracking control") 

The first part of the boolean function defines the type of vehicles used, while the second one defines the detection of the coordinated control techniques.

During the research process, a challenge faced was the limitations of various bibliographic databases, which are not fully equipped to accommodate this type of review. These limitations include restrictions on specific types of searches or searching within certain sections of papers. Consequently, the research using the boolean function was conducted within the titles, abstracts, and keywords of the documents.

Owing to these database constraints, as detailed in \cite{ierardi2018guidelines}, and considering the satisfactory coverage provided by the publishers as presented in Table \ref{fig: Databases_coverage}, the selected databases for this study were as follows:
\begin{itemize}
 \item IEEE Xplore Digital Library
 \item Scopus
 \item ACM Digital Library
 \item Science Direct
 \item Web of Science
\end{itemize}

\begin{table*}[H]
\caption[Databases coverage with respect to the content of the publishers]{Databases coverage with respect to the content of the publishers~\cite{ierardi2018guidelines}.}
	\centering
		\scalebox{.95}[0.95]{\begin{tabular}{cccccc}
		\toprule
		&\textbf{ IEEE Xplore }& \textbf{ACM Digital Library} & \textbf{Scopus} & \textbf{Web of Science} & \textbf{Science Direct}\\ 
		\noalign{\hrule height 0.5pt} 
		IEEE& \cellcolor{red!65} & \cellcolor{green!65} & \cellcolor{blue!65} & \cellcolor{orange!65} & \\ 
		IET& \cellcolor{red!65} & & & & \\ 
		Pegamon Elsevier& & \cellcolor{green!65} & \cellcolor{blue!65} & \cellcolor{orange!65} & \cellcolor{violet!65} \\ 
		Elsevier Science & & \cellcolor{green!65} & \cellcolor{blue!65} &\cellcolor{orange!65} & \cellcolor{violet!65} \\ 
		Wiley Blackwell & \cellcolor{red!65} & & \cellcolor{blue!65} &\cellcolor{orange!65} & \\ 
		Taylor and Francis & & \cellcolor{green!65} & \cellcolor{blue!65} & \cellcolor{orange!65}& \\ 
		Springer & &\cellcolor{green!65} & \cellcolor{blue!65} &\cellcolor{orange!65} & \\ 
		SIAM Publications & & \cellcolor{green!65}& \cellcolor{blue!65} & \cellcolor{orange!65}& \\ 
		Oxford University Press & & \cellcolor{green!65}& \cellcolor{blue!65} & \cellcolor{orange!65}& \\ 
		Korean Inst. Electrical Eng. & & & \cellcolor{blue!65} & & \\ 
		Sage Publications & &\cellcolor{green!65} &\cellcolor{blue!65} &\cellcolor{orange!65} & \\ 
		ASME & & & \cellcolor{blue!65}& & \\ 
		Microtome Publications & & \cellcolor{green!65}& & & \\ \noalign{\hrule height 1.0pt} 
	\end{tabular}}
	\label{fig: Databases_coverage}
\end{table*}

After identifying the research questions and creating the corresponding boolean function, it needs to be incorporated into various bibliographic databases, with adjustments made to align it with the search language specific to each database. In this study, the research was conducted by exclusively searching within the abstracts, titles, and keywords of the papers. This search yielded a total of 2334 items as of September 2023, as outlined in Table \ref{fig:database_res}.

\begin{table}[H]
\caption{Studies obtained by each database.}
	\centering
	\begin{tabular}{cc}\toprule
		{\textbf{Database}} & {\textbf{Studies}} \\ 
        \midrule
		Web of Science & 620 \\
		IEEE Xplore & 350\\
		ScienceDirect & 98 \\
		ACM Digital Library & 132 \\
		Scopus & 1134 \\
        \midrule
		& 2334   \\\bottomrule
	\end{tabular}
	\label{fig:database_res}
\end{table}

Another critical step within the planning phase is the establishment of inclusion and exclusion criteria. These criteria hold significant importance, serving as objective guidelines for the selection of studies eligible for inclusion in the review.
In order for a paper to be included in the review, it must meet all the inclusion criteria outlined in Table \ref{fig: Inclusion} and must not exhibit any of the exclusion criteria listed in Table \ref{fig: Exclusion}.
\begin{table}[H]
\caption{Inclusion criteria.}
	\centering
	\begin{tabular}{{>{\centering\arraybackslash}m{3in}}}\toprule
		\begin{itemize}
		 \item  Full paper available online (through search engines or by contacting the authors)
		\item  Use or propose a coordinated control technique of multiple ASVs
		\item  The system must have a dynamic model
		\end{itemize}\\
	\bottomrule
	\end{tabular}
	\label{fig: Inclusion}
\end{table}

\begin{table}[H]
\caption{Exclusion criteria.}
	\centering
	\begin{tabular}{{>{\centering\arraybackslash}m{3in}}}
	\toprule
		\begin{itemize}
		\item Secondary studies and gray literature
		\item Non-English written papers
		\item Duplicated studies 
		\item Fully actuated system
		\item Non-holonomic vehicles (without sway)
            \item Hybrid fleet (that includes underwater or aerial vehicles)
            \item Lack of tests or simulations
            \item Studies clearly irrelevant to the research
		\end{itemize}\\
	\bottomrule
	\end{tabular}
	\label{fig: Exclusion}
\end{table}
In addition to the various tools required to narrow down the initial pool of 2334 papers obtained through the boolean function search (refer to Table \ref{fig:database_res}), it is essential to establish guidelines for the selection of studies and the extraction of crucial data from each paper.

\subsection{Conducting the review} \label{subsec:conducting}
The main objectives during this phase include the following:
\begin{itemize}
    \item Significantly reducing the extensive pool of obtained studies through the application of inclusion and exclusion criteria.
    \item Creating a features table that highlights the principal characteristics of each article.
\end{itemize}

\begin{figure}
	\centering
		\includegraphics[scale=.6]{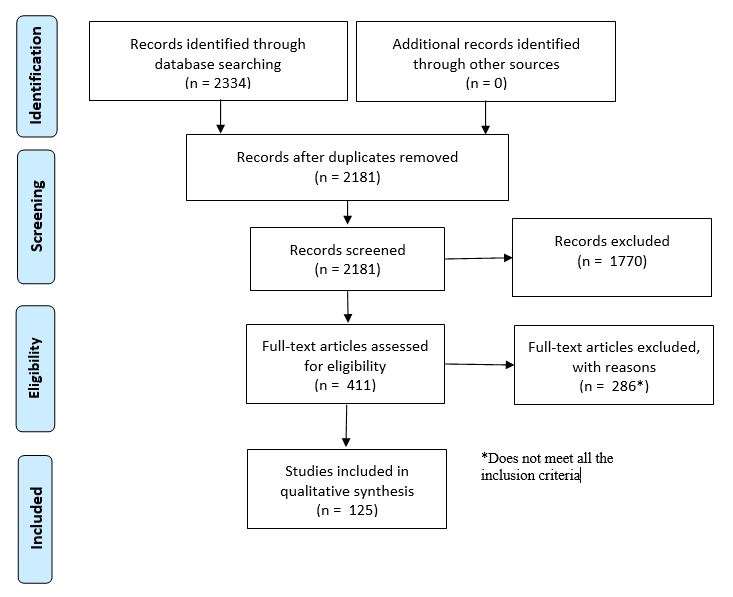}
	\caption{PRISMA flowchart.}
	\label{fig: prisma}
\end{figure}

\textbf{- Selection of primary studies:}
To facilitate the primary study selection, the PRISMA \cite{moher2009preferred} (Preferred Reporting Items for Systematic Reviews and Meta-Analyses) approach is frequently employed. This method includes 27 items and a flowchart designed to streamline and organize the entire process. As illustrated in Figure \ref{fig: prisma}, from the initial 2334 papers selected, the first screening removes duplicates, totaling 153 in this instance. This occurs because many databases overlap with certain publishers, as indicated in Table \ref{fig: Databases_coverage}. Utilizing software, such as Mendeley, is recommended for automatic duplicate detection.

The subsequent step involved reviewing the titles, abstracts, and keywords of the 2181 remaining papers, applying the set inclusion and exclusion criteria to them. Often, scrutinizing these specific sections alone is insufficient to ascertain full compliance with all criteria, necessitating a thorough read-through of the entire paper. This challenge is primarily attributed to the lack of standardized guidelines for crafting these sections in the engineering domain, unlike in other fields such as medicine or psychology. Following this process, out of the initial 2181 articles, 411 were selected for comprehensive review, and ultimately, only 125 papers fulfilled the criteria for inclusion in the study. These papers will be assessed to address the previously posed questions. The Conducting phase proved to be the most challenging and time-consuming part of the systematic review, given the meticulous analysis of 2181 papers. It is important to highlight that each paper undergoes an independent evaluation by two reviewers, and a primary study must gain approval from both to be included. In instances of disagreement between the reviewers, the responsibility of making the final decision falls to a third reviewer.

\textbf{- Study quality assessment:}
A quality assessment was conducted on the studies, meaning that all the criteria listed in Table \ref{fig: Quality_assessment} were examined. The outcomes of this quality evaluation are presented in Table \ref{fig: Quality_assessment_study}.

In addressing Question 2, a score of "Partially" has been allocated when the considered article does not sufficiently detail the communication topology.

Regarding Question 3, three major constraints have been primarily considered (each elaborately discussed in Section \ref{subsec:rq1.1}): the absence of a diagonal matrix, model uncertainties, and whether perturbations are considered. The evaluation criteria are established as follows: if all three constraints are met, the score assigned is "NO"; if none are met, the score is "HARD"; and if some are met, the score is "SOFT".

\begin{table}[H]
\caption{Checklist for quality assessment (\cite{ierardi2018guidelines}).}
\scriptsize{
	\centering
		\begin{tabular}{p{0.01\textwidth}p{0.25\textwidth}p{0.13\textwidth}}
			\toprule
		   & \textbf{Question} & \textbf{Score} \\ \midrule
        Q1 & Is the problem presented clearly? & Yes/Partially/No \\
        Q2 & Is the methodology used presented clearly? &  Yes/Partially/No \\
        Q3 & Are there any limitations and/or restrictions? &  Hard/Soft/No \\
        Q4 & Is there a discussion of the results? & Yes/Partially/No \\
        Q5 & Does it answer all the questions originally formulated by the SR? & Yes/Partially/No \\
        Q6 & Has it been cited by many authors?  & Cites/Year \\
        Q7 & Has it been published in a journal or conference proceeding?  & Journal/Conference \\ \bottomrule
	\end{tabular}}
\label{fig: Quality_assessment}
\end{table}

\textbf{- Extraction and synthesis of data:}
To compile and synthesize the data from the chosen papers, summary tables were developed and presented in Section \ref{sec:report}. These tables enumerate the most significant features of the papers pertinent to the research subject.
At this link (\url{https://hdl.handle.net/20.500.12412/5698}), an Excel document is available for readers, compiling all extracted features for each article into a single comprehensive table.

\begin{itemize}
    \item Year of publication.
    \item The sort of control technique used, such as adaptive, a model predictive, a sliding mode control or any other.
    \item If the effectiveness of the proposed control techniques has been demonstrated through simulations and/or experiments.
    \item The architecture of the control system is determined by whether it is implemented in a centralized or decentralized way.
    \item The information that needs to be exchanged between the ASVs: angle, position, speed and/or control action, as well as each vehicle regarding its own parameters.
    \item The strategies for guidance and control used to facilitate cooperative behaviors include leader-follower dynamics, flocking techniques, and graph-based approaches, among others.
    \item The various types of reference signals, such as trajectory-guided, path-guided, target-guided and formation stabilization coordinated controls.
    \item Motion control problems, if disturbances are considered and the fields of application in which these techniques are used.
    \item The communication protocol. We are interested in how each ASV communicates with the others. Specifically, we will note whether the control algorithm requires communication among all entities, only with adjacent ASVs, or exclusively with a leader, among other configurations.
    \item Advantages and/or limitations mentioned by the original study authors or identified by the reviewers are also considered. This aspect is the only subjective element. However, we believe it enables the inclusion of valuable insights that are not captured by the other data.
    \item If any machine learning techniques are used
\end{itemize}

\section{Report of the Systematic Review}\label{sec:report}
In recognition of its significance and breadth, the reporting segment of this review has been allocated its own dedicated section. In this section, we provide a clear and detailed answer to each of the research questions that were exposed at the beginning of the review.

\subsection{RQ.1: What coordinated control techniques are used in multiple autonomous surface vehicles?} \label{subsec:rq1}

In the world of ASVs, coordinated control techniques are pivotal for effectively managing and operating multiple vehicles synchronised. The variety of control methods employed in this domain reflects the complexity and diversity of tasks that ASVs are expected to perform. This review categorizes the control techniques into several distinct groups, each supported by numerous references that demonstrate their application in the field, as shown in Table \ref{fig: control_techniques}. 

 A quantitative analysis of this table reveals a large and diverse landscape of control techniques applied to ASV coordination, where the techniques aimed at maintaining the stability of the controllers (Lyapunov) with more than 70 different references and the techniques focused on adaptive control with more than 40 references stand out above all. This fact has its logic due to the need to correct uncertainties in mathematical models for real implementations. The use of non-linear models also adjusts to the fact that the most used techniques are focused on Lyapunov.
 \\

\begin{table}[htbp]
\caption{Checklist for quality assessment. Par= Partially, Jou=Journal, Con=Congress}
	\centering
        \scriptsize{
		\begin{tabular}{cccccccc}\toprule
                  \textbf{Ref}   & \textbf{Q1} & \textbf{Q2}  & \textbf{Q3 }& \textbf{Q4} & \textbf{Q5} & \textbf{Q6} & \textbf{Q7 }\\ \midrule
    \cite{Arrichiello201273} & Yes & Par & Hard & Yes & Par & 23/2012 & Jou \\
    \cite{BabiÄ‡2016361} & Yes & Par & Hard & Yes & Par & 07/2016 & Con \\
    \cite{Bai20205015} & Yes & Yes & Hard & Yes & Par & 01/2020 & Con \\
     \cite{bishop2012} & Yes & Par & No & Yes & Par & 07/2012 & Con \\
     \cite{BÃrhaug2011493} & Yes & Yes & Soft & Yes & Par & 203/2010 & Jou \\
     \cite{chen2019} & Yes & Yes & Soft & Yes & Yes & 53/2018 & Jou \\
     \cite{Chen2015} & Yes & Yes & Hard & Yes & Par & 67/2015 & Jou \\
     \cite{chen2022_hierarchical} & Yes & Yes & Hard & Yes & Par & 0/2022 & Con \\
     \cite{Chen2022} & Yes & Yes & No & Yes & Yes & 69/2022 & Jou \\
     \cite{chen2020} & Yes & Yes & Soft & Yes & Yes & 11/2020 & Jou \\
     \cite{Chen2022reinforcement} & Yes & Yes & Soft & Yes & Yes & 22/2022 & Jou \\
     \cite{Chen2015Global} & Yes & Yes & Soft & Yes & Par & 04/2014 & Con \\
     \cite{dai2022} & Yes & Par & No & Yes & Par & 88/2020 & Jou \\
     \cite{do2016} & Yes & Yes & Hard & Yes & Par & 103/2016 & Jou \\
     \cite{Do2012} & Yes & Yes & Soft & Yes & Par & 74/2011 & Jou \\
     \cite{dong2021} & Yes & Yes & Hard & Yes & Par & 07/2021 & Jou \\
     \cite{Dong2010} & Yes & Yes & Hard & Yes & Par & 61/2010 & Jou \\
     \cite{dong2022} & Yes & Yes & Soft & Yes & Par & 05/2022 & Jou \\
     \cite{Eek2021171} & Yes & Par & Soft & Yes & Par & 18/2021 & Jou \\
     \cite{fahimi2007SlidingMode} & Yes & Yes & Soft & Yes & Par & 226/2007 & Jou \\
     \cite{Fahimi2008} & Yes & Par & Hard & Yes & Par & 07/2008 & Jou \\
     \cite{Fahimi2007Nonlinear} & Yes & Yes & Soft & Yes & Par & 79/2007 & Jou \\
     \cite{Fan2017} & Yes & Yes & Hard & Yes & Par & 09/2017 & Con \\
     \cite{Fu2022} & Yes & Par & Hard & Yes & Par & 02/2022 & Jou \\
     \cite{Fu2018Bioinspired} & Yes & Yes & Hard & Yes & Par & 05/2018 & Jou \\
     \cite{Fu2018Formation} & Yes & Yes & Hard & Yes & Par & 13/2018 & Jou \\
     \cite{Gao2023Fixed} & Yes & Yes & Soft & Yes & Par & 36/2022 & Jou \\
     \cite{Gao2021} & Yes & Yes & Soft & Yes & Par & 48/2021 & Jou \\
     \cite{Ghommam2018} & Yes & Yes & Soft & Yes & Par & 126/2017 & Jou \\
     \cite{Ghommam2020} & Yes & Yes & No & Yes & Yes & 37/2020 & Jou \\
     \cite{Ghommam2009} & Yes & Yes & No & Yes & Par & 169/2009 & Jou \\
     \cite{Gu2019Antidisturbance} & Yes & Yes & Soft & Yes & Par & 72/2019 & Jou \\
     \cite{Gu2019Distributed} & Yes & Yes & Soft & Yes & Par & 73/2019 & Jou \\
     \cite{Gu2022} & Yes & Yes & Soft & Yes & Yes & 47/2021 & Jou \\
     \cite{Guo2021} & Yes & Yes & No & Yes & Par & 20/2021 & Jou \\
     \cite{Han2022} & Yes & Yes & Soft & Yes & Yes & 0/2022 & Con \\
     \cite{Hao2022} & Yes & Yes & Soft & Yes & Par & 01/2022 & Con \\
     \cite{He2021} & Yes & Par & No & Yes & Yes & 48/2021 & Jou \\
     \cite{He2022} & Yes & Yes & No & Yes & Yes & 29/2022 & Jou \\
     \cite{Hu2022} & Yes & Yes & Soft & Yes & Par & 32/2021 & Jou \\
     \cite{Hu2021} & Yes & Yes & Hard & Yes & Par & 40/2020 & Jou \\
     \cite{hu2023spontaneous} & Yes & Par & Hard & Yes & Par & 04/2023 & Jou \\
     \cite{hu2022bearing} & Yes & Yes & Hard & Yes & Par & 41/2021 & Jou \\
     \cite{Huang2021Adaptive} & Yes & Yes & Soft & Yes & Yes & 43/2021 & Jou \\
     \cite{Huang2019Improved} & Yes & Par & Soft & Yes & Yes & 71/2019 & Jou \\
     \cite{Huang2021Robust} & Yes & Yes & Soft & Yes & Par & 39/2021 & Jou \\
     \cite{Huang2022} & Yes & Yes & Soft & Yes & Par & 27/2022 & Jou \\
     \cite{Jiang2022NonfragileFormationControl} & Yes & Yes & No & Yes & Par & 01/2022 & Jou \\
     \cite{Jiang2022Sliding} & Yes & Yes & No & Yes & Par & 24/2022 & Jou \\
     \cite{Jiang2022Cooperative} & Yes & Yes & Soft & Yes & Par & 46/2021 & Jou \\
     \cite{Jiao2014} & Yes & Yes & Hard & Par & Par & 0/2014 & Jou \\
     \cite{Jin2022} & Yes & Yes & No & Yes & Yes & 10/2022 & Jou \\
     \cite{Jin2016} & Yes & Yes & No & Yes & Par & 419/2016 & Jou \\
     \cite{kim2021} & Yes & Yes & Hard & Yes & Par & 03/2021 & Jou \\
     \cite{Li2019} & Yes & Yes & Soft & Yes & Par & 03/2019 & Con \\
     \cite{Li2018Adaptive} & Yes & Yes & Soft & Yes & Par & 01/2017 & Con \\
     \cite{Li2018Finite} & Yes & Yes & Hard & Yes & Par & 297/2018 & Jou \\
     \cite{Li2020} & Yes & Par & Soft & Yes & Par & 06/2020 & Jou \\
     
        \bottomrule
	\end{tabular}}
\label{fig: Quality_assessment_study}
\end{table}

\begin{table}[H]
	\centering
        \scriptsize{
		\begin{tabular}{cccccccc}\toprule
                  \textbf{Ref}   & \textbf{Q1} & \textbf{Q2}  & \textbf{Q3 }& \textbf{Q4} & \textbf{Q5} & \textbf{Q6} & \textbf{Q7 }\\ \midrule
        \cite{Liang2019Novel} & Yes & Yes & Soft & Yes & Par & 60/2019 & Jou \\
     \cite{Liang2021SwarmVelocity} & Yes & Yes & Soft & Yes & Par & 18/2021 & Jou \\
     \cite{Liang2020Distributed} & Yes & Yes & Soft & Yes & Yes & 78/2020 & Jou \\
     \cite{Liang2020Adaptive} & Yes & Par & Soft & Yes & Par & 11/2020 & Jou \\
     \cite{Liang2022Secure} & Yes & Yes & No & Yes & Par & 16/2021 & Jou \\
     \cite{liang2019SwarmControl} & Yes & Par & Hard & Yes & Yes & 88/2019 & Jou \\
     \cite{Lin2022Neural} & Yes & Yes & No & Yes & Yes & 19/2022 & Jou \\
         \cite{Liu2016Coordinated} & Yes & Yes & Soft & Yes & Yes & 0/2016 & Con \\
         \cite{Liu2021RobustEvent} & Yes & Par & Soft & Yes & Yes & 02/2021 & Con \\
        \cite{Liu2020Collective}  & Yes & Yes & Hard & Yes & Par & 70/2019 & Jou \\
         \cite{Liu2017Saturated} & Yes & Yes & Soft & Yes & Par & 37/2017 & Jou \\
         \cite{Liu2022RobustFuzzy} & Yes & Yes & Soft & Yes & Yes & 01/2022 & Jou \\
         \cite{Liu2022Distributed} & Yes & Yes & Soft & Yes & Par & 15/2022 & Jou \\
         \cite{Liu2022Scanning} & Yes & Yes & Hard & Yes & Par & 52/2020 & Jou \\
        \cite{Liu2022Formation}  & Yes & Yes & Soft & Yes & Par & 42/2022 & Jou \\
        \cite{Lu2018}  & Yes & Yes & Soft & Yes & Yes & 45/2020 & Jou \\
        \cite{Lv2021}  & Yes & Yes & Soft & Yes & Par & 44/2021 & Jou \\
        \cite{Lv2023}  & Yes & Yes & Soft & Yes & Yes & 12/2022 & Jou \\
        \cite{Lv2022}  & Yes & Yes & Soft & Yes & Par & 40/2021 & Jou \\
        \cite{Ma2019}  & Yes & Yes & Hard & Yes & Yes & 01/2019 & Con \\
        \cite{Meng2012}  & Yes & Yes & Hard & Par & Par & 14/2012 & Con \\
        \cite{Mu2020}  & Yes & Yes & Soft & Yes & Yes & 13/2020 & Jou \\
        \cite{Ning2022}  & Yes & Yes & Soft & Yes & Yes & 14/2022 & Jou \\
        \cite{Panagou2013}  & Yes & Yes & Hard & Yes & Par & 14/2013 & Con \\
        \cite{Park2022}  & Yes & Yes & No & Yes & Yes & 15/2022 & Jou \\
        \cite{Park2021}  & Yes & Yes & No & Yes & Par & 66/2021 & Jou \\
        \cite{Peng2012NeuralAdaptive}  & Yes & Yes & Soft & Yes & Yes & 08/2012 & Con \\
        \cite{Peng2012RobustLeader}  & Yes & Yes & No & Yes & Yes & 176/2011 & Jou \\
        \cite{Peng2013AdaptiveDynamic}  & Yes & Yes & No & Yes & Yes & 498/2012 & Jou \\
        \cite{Peng2012RobustLeader}  & Yes & Yes & No & Yes & Yes & 29/2012 & Jou \\
        \cite{Peng2019PathGuided}  & Yes & Yes & Soft & Yes & Yes & 103/2019 & Jou \\
        \cite{Qin2017}  & Yes & Yes & Soft & Yes & Par & 44/2017 & Jou \\
        \cite{Riahifard2020}  & Yes & Yes & Soft & Yes & Par & 11/2019 & Jou \\
        \cite{Ringback2021}  & Yes & Yes & Hard & Yes & Par & 35/2020 & Jou \\
        \cite{Shojaei2015}  & Yes & Yes & Soft & Yes & Yes & 166/2015 & Jou \\
        \cite{Song2022}  & Yes & Yes & Soft & Yes & Yes & 04/2022 & Jou \\
        \cite{Sun2020}  & Yes & Yes & Hard & Yes & Par & 43/2020 & Jou \\
        \cite{Sun2018}  & Yes & Yes & Soft & Yes & Par & 150/2018 & Jou \\
        \cite{Sun2022}  & Yes & Par & Soft & Yes & Yes & 08/2022 & Jou \\
        \cite{Tang2023}  & Yes & Yes & Hard & Yes & Par & 19/2022 & Jou \\
        \cite{Wang2022}  & Yes & Yes & Soft & Yes & Par & 13/2022 & Jou \\
        \cite{Wang2019}  & Yes & Yes & Soft & Yes & Yes & 20/2019 & Jou \\
        \cite{Wang2021}  & Yes & Yes & Soft & Yes & Par & 29/2021 & Jou \\
        \cite{Wang2012}  & Yes & Yes & No & Yes & Yes & 09/2012 & Con \\
        \cite{Wu2022}  & Yes & Yes & Soft & Yes & Yes & 40/2022 & Jou \\
        \cite{Wu2021}  & Yes & Yes & Soft & Yes & Yes & 07/2021 & Jou \\
        \cite{Wu2023}  & Yes & Yes & Soft & Yes & Yes & 11/2022 & Jou \\
        \cite{Xia2021}  & Yes & Yes & Soft & Yes & Yes & 11/2022 & Jou \\
        \cite{Xia2022}  & Yes & Yes & Soft & Yes & Par & 10/2022 & Jou \\
        \cite{Xia2020}  & Yes & Yes & Soft & Yes & Par & 18/2020 & Jou \\
        \cite{Xiong2022}  & Yes & Yes & No & Yes & Yes & 03/2022 & Jou \\
        \cite{Ye2021}  & Yes & Par & Hard & Yes & Par & 02/2021 & Jou \\
        \cite{Yu2019}  & Yes & Yes & Soft & Yes & Par & 27/2019 & Jou \\
        \cite{Yue2019}  & Yes & Yes & Hard & Yes & Par & 01/2019 & Jou \\
        \cite{Yuhan2022}  & Yes & Yes & Soft & Yes & Par & 05/2022 & Jou \\
        \cite{Zhang2022EventTriggered}  & Yes & Yes & Soft & Yes & Yes & 33/2022 & Jou \\
        \cite{Zhang2020FormationControlMultiple}  & Yes & Par & Hard & Yes & Par & 04/2020 & Jou \\
        \cite{Zhang2021LVS}  & Yes & Yes & Soft & Yes & Yes & 22/2021 & Jou \\
        \cite{Zhang2022Bearing} & Yes & Yes & Soft & Yes & Yes & 07/2022 & Jou \\
        \cite{Zhang2021NovelEventTriggered} & Yes & Yes & Soft & Yes & Yes & 17/2021 & Jou \\
        \cite{Zhang2022AdaptiveDistributed} & Yes & Yes & Soft & Yes & Yes & 10/2022 & Jou \\
        \cite{Zhang2022RobustAdaptive} & Yes & Yes & Soft & Yes & Yes & 05/2022 & Jou \\
        \cite{Zhang2019FormationControlAutonomous} & Yes & Yes & Hard & Yes & Par & 03/2019 & Con \\
        \cite{Zhao2021} & Yes & Yes & Soft & Yes & Yes & 93/2021 & Jou \\
        \cite{Zhou2020} & Yes & Yes & Hard & Yes & Par & 108/2020 & Jou \\
        \cite{Zhou2022} & Yes & Yes & Soft & Yes & Yes & 22/2022 & Jou \\
        \cite{Zhuang2019} & Yes & Par & Hard & Yes & Par & 18/2019 & Jou \\
        \bottomrule
	\end{tabular}}
\end{table}

Among the commonly used techniques, controllers based on back-stepping, neural networks and radial basis functions stand out. These trends are explained above all due to the non-linearities of the ASV models, as well as the rise of machine learning techniques. At a more moderate level, techniques based on sliding modes, robust control, predictive control and those based on extended-state-observer stand out.

\begin{figure}
	\centering
		\includegraphics[trim={0cm} {0cm} {0cm} {0cm},clip,width=0.5\textwidth]{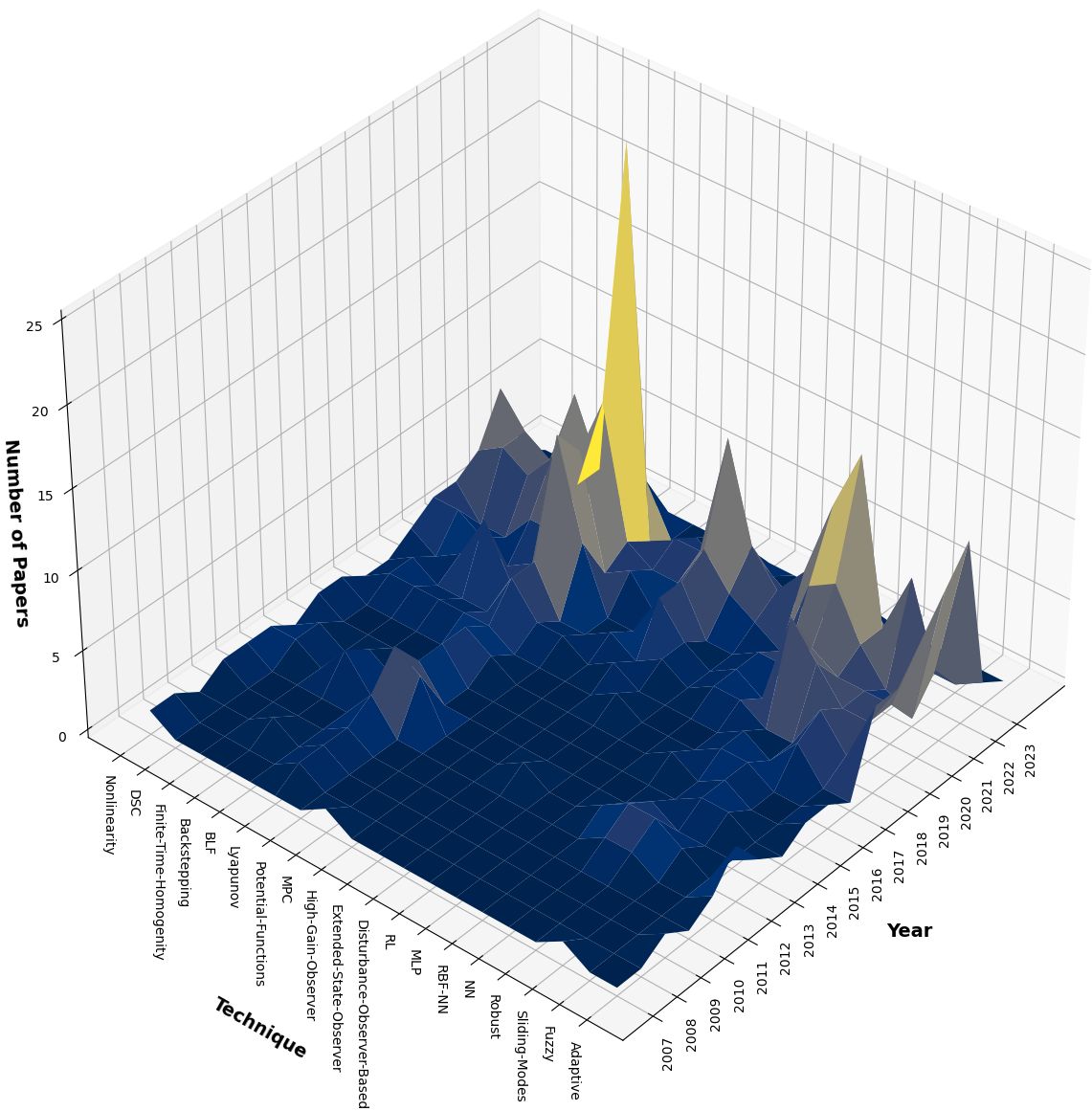}
	\caption{Development of Control Techniques, 2007-2023}
	\label{FIG:percentage_controls}
\end{figure}

In Figure \ref{FIG:percentage_controls}, trends in the research community can be seen regarding the development of specific control techniques in recent years, such as Lyapunov, extended-state-observer, NN, robust control and adaptive control. The analysis also points to possible synergies between hybrid control strategies that leverage the strengths of multiple approaches. 

\begin{table*}[H]
\caption{Comparison of control techniques and their associated references.}
	\centering
 \scriptsize{
		\begin{tabular}{cl}\toprule
                  Control technique   & References \\ \midrule
Adaptive &  \citep{BabiÄ‡2016361}, \citep{chen2019}, \citep{Chen2022}, \citep{chen2020}, \citep{dai2022}, \citep{Eek2021171}, \citep{Fu2018Bioinspired}, \citep{Ghommam2018},  \citep{Ghommam2009}, \citep{Gu2022}, \cite{He2021}, \citep{He2022}, \citep{Han2022}, \citep{Hu2021}, \citep{Huang2021Adaptive}, \citep{Huang2019Improved}, \citep{Huang2021Robust}, \citep{Li2018Adaptive}, \citep{Li2020}, \\
&  \citep{Liang2020Adaptive}, \citep{Liu2021RobustEvent}, \citep{Lu2018}, \citep{Lv2023}, \citep{Ma2019}, \citep{Mu2020}, \citep{Ning2022},  \citep{Park2022}, \citep{Peng2012NeuralAdaptive}, \citep{Peng2011RobustAdaptive}, \citep{Peng2013AdaptiveDynamic}, \citep{Peng2019PathGuided}, \citep{Peng2012RobustLeader}, \citep{Song2022}, \citep{Sun2020}, \citep{Sun2018}, \citep{Riahifard2020}, \citep{Shojaei2015},   \\
& \citep{Wang2012}, \citep{Wu2021}, \citep{Yuhan2022},  \citep{Zhang2022EventTriggered}, \citep{Zhang2021LVS}, \citep{Zhang2021NovelEventTriggered}, \citep{Zhang2022AdaptiveDistributed}, \citep{Zhang2022RobustAdaptive}, \citep{Zhou2020}, \citep{Zhou2022}, \citep{Wang2019} \\
& \\
Fuzzy &  \citep{Jiang2022Cooperative}, \citep{Li2018Adaptive}, \citep{Liang2019Novel}, \citep{Liang2020Adaptive}, \citep{Liu2022RobustFuzzy}, \citep{Zhang2019FormationControlAutonomous}, \citep{Zhou2020}\\
& \\
Sliding-modes &  \citep{Bai20205015}, \citep{chen2020}, \citep{Chen2022reinforcement}, \citep{dong2022}, \citep{Eek2021171}, \citep{fahimi2007SlidingMode}, \citep{Fu2018Bioinspired}, \citep{Jiang2022Sliding}, \citep{Jiao2014}, \citep{Li2018Finite}, \citep{Liang2019Novel}, \citep{Liu2022Formation}, \citep{Meng2012}, \citep{Qin2017}, \citep{Song2022} \\
&   \citep{Sun2018}, \citep{Sun2022}, \citep{Zhuang2019}\\
& \\
Robust & \citep{Ghommam2020}, \citep{Huang2021Robust}, \citep{Jiao2014}, \citep{Liang2019Novel}, \citep{liang2019SwarmControl}, \citep{Liu2021RobustEvent}, \citep{Liu2022RobustFuzzy}, \citep{Park2021}, \citep{Peng2011RobustAdaptive}, \citep{Peng2012RobustLeader}, \citep{Wang2022}, \citep{Yue2019}, \citep{Peng2019PathGuided}, \citep{Riahifard2020}, \citep{Shojaei2015}, \citep{Wang2012}, \citep{Yuhan2022}\\
& \\
PID & \citep{kim2021}, \citep{Liu2020Collective}, \citep{Ringback2021}\\
& \\
Extended-state-observer-based  & \citep{Fu2022}, \citep{Gao2021}, \citep{Gao2023Fixed}, \citep{Ghommam2020}, \citep{Gu2019Antidisturbance}, \citep{Gu2019Distributed}, \citep{Guo2021}, \citep{Hu2022}, \citep{Hu2021}, \citep{Jiang2022Cooperative}, \citep{Liu2017Saturated}, \citep{Lv2021}, \citep{Lv2022}, \citep{Ning2022}, \citep{Xia2022}, \citep{Yu2019}, \\
&  \citep{Wu2022}, \citep{Wu2023}, \citep{Xia2020}, \citep{Zhang2022RobustAdaptive} \\
& \\
Disturbance-observer-based & \citep{do2016}, \citep{Do2012}, \citep{dong2022}, \citep{Huang2019Improved},  \citep{Wu2021}, \citep{Zhou2022}\\
& \\
Nonlinearity  &  \citep{BÃrhaug2011493}, \citep{Chen2015Global}, \citep{dong2021}, \citep{Dong2010}, \citep{Fahimi2008}, \citep{Fahimi2007Nonlinear}, \citep{Gao2023Fixed}, \citep{Liang2019Novel}, \citep{Liang2022Secure}, \citep{Liu2022Distributed}, \citep{Meng2012}, \citep{Zhang2020FormationControlMultiple}, \\
& \citep{Sun2022}, \citep{Xia2022}, \citep{Zhang2022EventTriggered}, \citep{Zhang2021LVS}, \citep{Zhao2021}, \citep{Zhou2020}, \citep{Zhou2022}, \citep{Zhuang2019}\\
& \\
Model predictive &  \citep{Fahimi2007Nonlinear}, \citep{Fan2017}, \citep{Lv2021}, \citep{Lv2023}, \citep{Liu2022Distributed}, \citep{Liu2022Scanning}, \citep{Sun2020},\\
& \\
Redundant manipulator & \citep{bishop2012}\\
& \\
Lyapunov &  \citep{Bai20205015}, \citep{BÃrhaug2011493}, \citep{Chen2015}, \citep{chen2020}, \citep{Chen2022reinforcement}, \citep{Dong2010}, \citep{dong2022},  \citep{Eek2021171}, \citep{Fahimi2008}, \citep{Fu2022},  \citep{Gao2023Fixed},  \citep{Gao2021}, \citep{Ghommam2020},  \citep{Ghommam2009}, \citep{Gu2019Antidisturbance}, \citep{Gu2019Distributed}, \citep{Gu2022}, \citep{Han2022},\\
& \citep{Hao2022}, \citep{He2022}, \citep{Hu2021}, \citep{Hu2022}, \citep{hu2022bearing},\citep{hu2023spontaneous}, \citep{Huang2021Adaptive}, \citep{Huang2019Improved}, \citep{Huang2021Robust}, \citep{Huang2022},\citep{Jiang2022NonfragileFormationControl}, \citep{Jiao2014}, \citep{Li2018Adaptive}, \citep{Li2019}, \citep{Liang2019Novel}, \citep{Li2020}, \citep{Liang2020Distributed}, \citep{liang2019SwarmControl},  \\
& \citep{Liu2022Scanning}, \citep{Liu2022Formation}, \citep{Ma2019}, \citep{Meng2012}, \citep{Mu2020}, \citep{Ning2022}, \citep{Panagou2013}, \citep{Park2022}, \citep{Peng2012NeuralAdaptive}, \citep{Peng2011RobustAdaptive}, \citep{Peng2012RobustLeader}, \citep{Park2021}, \citep{Tang2023}, \citep{Zhang2022Bearing}, \citep{Lu2018}, \citep{Lv2022},     \\
& \citep{Peng2013AdaptiveDynamic}, \citep{Peng2019PathGuided}, \citep{Qin2017}, \citep{Riahifard2020}, \citep{Ringback2021},  \citep{Shojaei2015}, \citep{Song2022}, \citep{Sun2018}, \citep{Sun2022}, \citep{Wang2022}, \citep{Wang2012}, \citep{Wu2021}, \citep{Xia2021}, \citep{Xia2022}, \citep{Yue2019}, \\
& \citep{Zhang2022EventTriggered},  \citep{Zhang2021NovelEventTriggered}, \citep{Zhang2022AdaptiveDistributed}, \citep{Zhang2022RobustAdaptive}, \citep{Zhou2020}, \citep{Zhou2022}, \citep{Zhuang2019}\\
& \\
Back-stepping & \citep{Chen2022}, \citep{chen2022_hierarchical}, \citep{dai2022}, \citep{do2016}, \citep{Do2012}, \citep{dong2021}, \citep{dong2022}, \citep{Fu2018Formation}, \citep{Fu2018Bioinspired}, \citep{Ghommam2018}, \citep{Ghommam2009}, \citep{Guo2021}, \citep{He2021}, \citep{Huang2022}, \citep{Li2018Adaptive}, \citep{Liang2020Adaptive}, \citep{Panagou2013}, \\
& \citep{Peng2012RobustLeader}, \citep{Xia2021}, \citep{Wu2021}, \citep{Xia2022}, \citep{Yuhan2022}, \citep{Zhang2022EventTriggered}, \citep{Zhang2020FormationControlMultiple}\\
& \\
Finite-time-homogeneity  &  \citep{Fu2022}, \citep{Ye2021}, \citep{Han2022}, \citep{Huang2021Adaptive}, \citep{Huang2019Improved}, \citep{Jin2016}, \citep{Liang2021SwarmVelocity}, \citep{Lin2022Neural},  \citep{Wu2023}\\
& \\
Barrier Lyapunov function  &  \citep{dai2022}, \citep{Ghommam2018}, \citep{Guo2021}, \citep{He2021}, \citep{Jin2016}, \citep{Liang2021SwarmVelocity}, \citep{Liang2020Adaptive}, \citep{Lin2022Neural}, \citep{Wang2019}, \citep{Wang2021}, \citep{Wu2022}, \citep{Wu2023}, \citep{Xia2020}, \citep{Xiong2022}, \citep{Wu2022},  \citep{Wu2023}\\
& \\
Dynamic surface & \citep{chen2019}, \citep{Liu2016Coordinated}, \citep{dong2022}, \citep{Gu2019Antidisturbance}, \citep{Han2022}, \citep{He2021}, \citep{Huang2021Robust}, \citep{Liu2017Saturated}, \citep{Liu2022RobustFuzzy}, \citep{Peng2013AdaptiveDynamic}, \citep{Wang2012}, \citep{Wu2021}, \citep{Xiong2022}, \citep{Zhang2021NovelEventTriggered} \\
& \\
Reinforcement learning & \citep{Chen2022reinforcement}, \citep{Jin2022}, \citep{Zhao2021}\\
& \\
Neural-network   & \citep{chen2019}, \citep{Chen2022}, \citep{Chen2022reinforcement}, \citep{Ghommam2020}, \citep{He2021}, \citep{He2022}, \citep{Huang2021Adaptive},  \citep{Huang2019Improved}, \citep{Liang2020Distributed}, \citep{Liu2016Coordinated}, \citep{Liu2017Saturated}, \citep{Lu2018}, \citep{liang2019SwarmControl}, \citep{Lin2022Neural}, \citep{Liu2021RobustEvent}, \citep{Liu2022RobustFuzzy}, \citep{Lv2023}, \citep{Ma2019}, \\
& \citep{Mu2020}, \citep{Park2022}, \citep{Peng2011RobustAdaptive}, \citep{Peng2013AdaptiveDynamic}, \citep{Peng2012RobustLeader}, \citep{Zhao2021}, \citep{Peng2019PathGuided}, \citep{Sun2022}, \citep{Wang2019}, \citep{Wang2012}, \citep{Wu2022},  \citep{Wu2023}, \citep{Xia2021}, \citep{Zhang2022EventTriggered}, \citep{Zhang2021LVS}, \citep{Zhang2022Bearing}, \\
& \citep{Zhang2021NovelEventTriggered}, \citep{Zhang2022AdaptiveDistributed}, \citep{Zhang2022RobustAdaptive}, \citep{Zhou2022}\\
& \\
Potential functions & \citep{chen2019}, \citep{Chen2015}, \citep{Do2012}, \citep{Ghommam2020}, \citep{Liang2019Novel}, \citep{Liang2021SwarmVelocity}, \citep{Liang2020Distributed}, \citep{liang2019SwarmControl}, \citep{Peng2012NeuralAdaptive}, \citep{Peng2019PathGuided}, \citep{Qin2017}, \citep{Ringback2021}, \citep{Sun2020}, \citep{Sun2022}, \citep{Xia2021}, \citep{Xia2022}, \citep{Zhuang2019}\\
& \\
Null-Space-based Behavioral    &  \citep{Arrichiello201273}, \citep{Eek2021171}\\
& \\
Radial basis function & \citep{chen2020}, \citep{Gu2022}, \citep{Han2022}, \citep{He2022}, \citep{Huang2021Adaptive}, \citep{Liu2021RobustEvent}, \citep{Lu2018}, \citep{Ning2022}, \citep{Peng2012NeuralAdaptive}, \citep{Shojaei2015}, \citep{Song2022}, \citep{Wu2021}, \citep{Xia2021}, \citep{Xiong2022}, \citep{Zhang2022EventTriggered}, \citep{Zhang2022Bearing},\\
& \citep{Zhang2021NovelEventTriggered}, \citep{Zhang2022RobustAdaptive}\\
& \\
Minimal learning parameter & \citep{Han2022}, \citep{Huang2019Improved}, \citep{Liu2022RobustFuzzy}, \citep{Lu2018}, \citep{Mu2020}, \citep{Song2022}, \citep{Zhang2022RobustAdaptive} \\
& \\
High-gain-observer &\citep{dai2022}, \citep{Huang2022}, \citep{Lin2022Neural}, \citep{Lu2018},\citep{Ghommam2018}, \citep{Wang2021} \\
& \\
Flexible Formation Tracking & \citep{Tang2023}\\
& \\
Deep learning & \citep{Chen2022reinforcement} ,\citep{Xia2021}, \citep{Zhang2022AdaptiveDistributed}, \citep{Zhang2022RobustAdaptive}, \citep{Zhao2021}\\
& \\
$H_{\infty}$ & \citep{Yue2019}\\
\bottomrule
	\end{tabular}}
\label{fig: control_techniques}
\end{table*}

\subsection{RQ.1.1: What are the limitations and advantages of the different techniques?} \label{subsec:rq1.1}

\subsubsection{Non-diagonal inertia matrices}\label{subsubsec:non-diagonal}
The main difficulty in controlling underactuated ships is that they are not actuated in the sway axis \citep{do2005global}.
In the majority of the articles reviewed, about $80\%$ the mass and damping matrices for ships are presumed to be diagonal. Such restrictive assumptions would necessitate the ship being akin to a semi-submerged sphere \citep{do2005global}. However, in practical terms, these assumptions do not entirely apply to real ships. This is evidenced by the fact that only 2 out of 17 experimental studies accounted for this limitation (see Table \ref{fig: limitation}).

For this reason, non-diagonal inertia matrices in ASVs present a nuanced challenge in control system design. Diagonality in these matrices typically allows for the decoupling of specific dynamics from others, significantly simplifying the development of control systems. However, when this condition is not met, it tends to exacerbate existing issues, particularly with yaw control. 

Indeed, when the inertia matrix is diagonal, sway and yaw dynamics are decoupled, aiding the yaw controller design to some extent. Yet, simulations have shown that this decoupling does not entirely mitigate the problem of sway generation during turns, a phenomenon that remains uncontrollable. The work of Do and Pan \citep{do2005global}, stands out for its thorough explanation of this issue and offers a potential solution, according to the literature and simulations conducted thus far.
In fact, many authors of the aforementioned percentage papers recognized difficulties caused by the nonzero off-diagonal terms of system matrices. They neglected these terms and left the topic of dealing with these terms for a future work.



\subsubsection{Consideration of uncertainties}\label{subsubsec:uncertainties}

In the realm of ASVs, the ability to maintain coordinated control in the face of environmental disturbances is paramount. Below it delves into the various types of disturbances that can affect the performance and operational effectiveness of ASVs participating in cooperative tasks. Disturbances, by nature, introduce unpredictability and complexity to the control systems of ASVs, demanding robust and adaptive control strategies to mitigate their effects.
Environmental factors such as wind, waves, and currents pose significant challenges to maintaining formation integrity, trajectory tracking, and overall mission success. In Table \ref{fig: disturbance}, it is highlighted which articles consider certain factors as disturbances and which do not. Among the distinct types of disturbances considered are:
\begin{itemize}
    \item General disturbances encompass any external, nonspecific factors that may affect the behavior or trajectory of autonomous vehicles, such as environmental variability, modeling errors, or inaccuracies in sensors and actuators. 
    \item Wind disturbances are aerodynamic forces and moments generated by air movement that can alter the direction, speed, and stability of surface autonomous vehicles. 
    \item Wave disturbances result from changes in the water surface, usually caused by wind, that can affect the navigation and stability of marine vehicles, inducing unwanted movements and trajectory variations. 
    \item Current disturbances refer to variations in the underlying water movement that can influence the speed and direction of autonomous surface vehicles, requiring control adjustments to maintain a desired trajectory.
    \item Connection disturbances refer to operational disruptions stemming from physical connections, such as tethers or cables, between vehicles. These disturbances are pivotal in scenarios where ASVs are linked, as they can significantly affect formation dynamics and the execution of coordinated tasks.
    \item None, if none of the aforementioned disturbances are considered.
    
\end{itemize}
The presence of studies that concurrently consider wind, wave, and current disturbances highlights the importance of addressing these factors in an integrated manner, given their interplay and combined impact on vehicle behavior. This holistic approach is crucial for developing more robust and adaptable control systems capable of handling the complexities and dynamics of the marine environment. Papers with this feature are \citep{Ghommam2009}, \citep{Gu2019Distributed}, \citep{Huang2021Adaptive}, \citep{Liu2017Saturated}, \citep{Shojaei2015},\citep{Wu2021}, \citep{Wu2022}, \citep{Wu2023}, \citep{Zhang2021LVS}, \citep{Zhang2022Bearing}. The wide array of references considering general disturbances indicates a common approach in the literature to evaluate the robustness and effectiveness of control algorithms under varied and unspecified perturbations. It is worth mentioning that only one article was found in the category of connection disturbances \citep{Arrichiello201273}, where the two vehicles are connected by a floating flexible rope.

\begin{table*}[H]
\caption{Disturbances considered}
	\centering
		\begin{tabular}{cc}\toprule
                  Disturbance   & References \\ \midrule
 General&  \citep{Bai20205015}, \citep{bishop2012}, \citep{chen2019}, \citep{Chen2022}, \citep{chen2020}, \citep{Chen2022reinforcement}, \citep{dai2022}, \citep{do2016}, \citep{dong2021}, \citep{dong2022}, \citep{Fu2022}, \citep{Fu2018Bioinspired}, \citep{Fu2018Formation}, \citep{Gao2023Fixed}, \citep{Gao2021}, \citep{Ghommam2018}, \citep{Ghommam2020}, \citep{Gu2019Antidisturbance}, \citep{Gu2022}, \citep{Guo2021}, \citep{Han2022}, \citep{hu2023spontaneous}\\
 & \citep{Hao2022}, \citep{He2021}, \citep{He2022}, \citep{Hu2022}, \citep{Huang2022}, \citep{Jiang2022NonfragileFormationControl}, \citep{Jiang2022Sliding}, \citep{Jiang2022Cooperative}, \citep{Jiao2014}, \citep{Jin2022}, \citep{Jin2016}, \citep{Li2019}, \citep{Li2018Adaptive}, \citep{Li2018Finite}, \citep{Li2020}, \citep{Liang2019Novel}, \citep{Liang2021SwarmVelocity}, \citep{Liang2020Distributed}, \citep{Liang2020Adaptive}, \citep{Liang2022Secure}, \citep{liang2019SwarmControl}\\
 & \citep{Lin2022Neural}, \citep{Liu2020Collective}, \citep{Liu2022RobustFuzzy}, \citep{Liu2022Distributed}, \citep{Liu2022Formation}, \citep{Lu2018}, \citep{Lv2021}, \citep{Lv2023}, \citep{Lv2022}, \citep{Ma2019}, \citep{Meng2012}, \citep{Mu2020}, \citep{Ning2022}, \citep{Panagou2013}, \citep{Park2022}, \citep{Park2021}, \citep{Peng2012NeuralAdaptive}, \citep{Peng2011RobustAdaptive}, \citep{Peng2013AdaptiveDynamic}, \citep{Peng2012RobustLeader}, \citep{Peng2019PathGuided}, \citep{Qin2017}\\
 & \citep{Riahifard2020}, \citep{Ringback2021}, \citep{Song2022}, \citep{Sun2018}, \citep{Sun2022}, \citep{Tang2023}, \citep{Wang2022}, \citep{Wang2019}, \citep{Wang2021}, \citep{Wang2012}, \citep{Xia2021}, \citep{Xia2022}, \citep{Xia2020}, \citep{Xiong2022}, \citep{Yu2019}, \citep{Yuhan2022}, \citep{Zhao2021}\\
 &\\
 Wind & \citep{Do2012}, \citep{Ghommam2009}, \citep{Gu2019Distributed}, \citep{Huang2021Adaptive},\citep{Huang2021Robust}, \citep{Liu2017Saturated}, \citep{Shojaei2015}, \citep{Wu2021}, \citep{Wu2022}, \citep{Wu2023}, \citep{Zhang2022EventTriggered}, \citep{Zhang2021LVS}, \citep{Zhang2022Bearing}, \citep{Zhang2021NovelEventTriggered}, \citep{Zhang2022AdaptiveDistributed}, \citep{Zhang2022RobustAdaptive} \\
  &\\
 Wave &  \citep{BÃrhaug2011493}, \citep{fahimi2007SlidingMode}, \citep{Fahimi2008}, \citep{Fahimi2007Nonlinear}, \citep{Ghommam2009}, \citep{Gu2019Distributed}, \citep{Huang2021Adaptive}, \citep{Huang2019Improved}, \citep{Huang2021Robust}, \citep{Liu2016Coordinated}, \citep{Liu2021RobustEvent}, \citep{Liu2017Saturated}, \citep{Shojaei2015}, \citep{Wu2021}, \citep{Wu2022}, \citep{Wu2023}, \citep{Zhang2022EventTriggered}, \citep{Zhang2021LVS}, \citep{Zhang2022Bearing}, \citep{Zhang2021NovelEventTriggered}, \\
 & \citep{Zhang2022AdaptiveDistributed}, \citep{Zhang2022RobustAdaptive}, \citep{Zhou2022}  \\
  &\\
 Current &  \citep{BabiÄ‡2016361}, \citep{Do2012}, \citep{Eek2021171}, \citep{Ghommam2009}, \citep{Gu2019Distributed}, \citep{Huang2021Adaptive}, \citep{Liu2017Saturated}, \citep{Liu2022Scanning}, \citep{Shojaei2015}, \citep{Wu2021}, \citep{Wu2022}, \citep{Wu2023}, \citep{Zhang2021LVS}, \citep{Zhang2022Bearing}, \citep{Zhou2022} \\
  &\\
 Connection & \citep{Arrichiello201273}\\
  &\\
 None & \citep{Chen2015}, \citep{chen2022_hierarchical}, \citep{Chen2015Global}, \citep{Dong2010}, \citep{Fan2017}, \citep{Hu2021}, \citep{hu2022bearing}, \citep{kim2021}, \citep{Sun2020}, \citep{Ye2021}, \citep{Yue2019}, \citep{Zhang2020FormationControlMultiple}, \citep{Zhang2019FormationControlAutonomous}, \citep{Zhou2020}, \citep{Zhuang2019}\\
\bottomrule
	\end{tabular}
\label{fig: disturbance}
\end{table*}

\subsubsection{Constrained variables} \label{subsubsec:Constrained}



The explicit consideration of the constraints of the physical variables in the ASVs is frequently important to avoid problems in the control loops, damages to the hardware and undesirable dynamic behaviours at the ASV or at the whole fleet level. 

Particularly important is to take into account the limitations of the ASVs actuators, as problems like windup or bang-bang behaviour are easily met when these constraints are not addressed in the design of the controllers.

Next, we summarize the main techniques found useful to deal with these constraints:

\begin{itemize}
    \item Tangent hyperbolic function: the $tanh$ function can be used to map the computed control signals to the range within the saturated values of the actuator. This map is smooth and provides a gradual and continuous transition near the saturation limits, which makes it easier both to analyze and guarantee the system stability using Lyapunov theory and to avoid abrupt changes in the control signals preventing overshoots or oscillations. This technique have been applied in conjunction with adaptive controllers (\citep{Riahifard2020} and \cite{Shojaei2015}), barrier Lyapunov function (BLF)-based controllers \citep{Xia2020}, backstepping (\citep{Xia2021} and \cite{BÃrhaug2011493}), and sliding-mode control \citep{Chen2022reinforcement}.
    \item Auxiliary systems: the inclusion of auxiliary systems in the control design also prevents undesirable closed-loop dynamical behaviours and yields to a smooth mapping of the control signals that makes it possible to use Lyapunov stability analysis. In this case, the reference or error signals are modified according to the dynamics of an auxiliary dynamical system, ensuring that the control input reaches saturated values smoothly and does not exceed them. Auxiliary systems have been used with adaptive controllers in \citep{Mu2020} and \citep{Zhou2020}, BLFs, and backstepping control in \citep{Xia2022}, \citep{Huang2022}.
    \item Explicit constraints: the Model predictive control  (MPC) techniques in Table \ref{fig: limitation} naturally address constrained variables introducing them directly in the set of constraints of the optimization problem that computes the optimal control inputs. These constrains may include input saturation and also saturations in the input change rate. Some BLF formulations can also explicitly include constraints. In particular, in \citep{Wu2022} and \citep{Wu2023}, 
    quadratic optimization problems are formulated based on control BLF to find the optimal control inputs taking into account stability constraints, safety constraints and input constraints, and the problem is solved using a recurrent neural network.  Differently, \citep{Xiong2022} considers constraints in the velocities, and feedback mechanism is proposed to revise the trajectory and velocity of leaders when followers encounter velocity saturation.
    \item Other techniques: In \cite{Yuhan2022}, a nonlinear function capturing saturations in the input is introduced in conjunction with an adaptive controller to avoid input overload, although the effectiveness of this technique is unclear. Moreover, \citep{Fu2018Bioinspired} introduces a bio-inspired velocity controller that avoid saturations in the actual low-level controller that manages the actuators. In \citep{Liu2017Saturated}, a kinetic control law based on back stepping and a projection operator makes it possible to upper bound the control inputs a priori by design. Finally, \citep{bishop2012} considers limitations in the surge velocities and the yaw rate to accommodate the hull speed and the ship's turning radius, respectively, modifying desired velocities to match these constraints.
    
\end{itemize}

\begin{table*}[h]
\caption{Limitations and advantages of the different techniques for each article}
	\centering
	\scriptsize{
        \begin{tabular}{{>{\centering\arraybackslash}m{2.165em}
                >{\centering\arraybackslash}m{22.15em}
                ccccccccccc}}\toprule
        \multicolumn{13}{c}{\textbf{Simulation}} \\ \hline
                Ref    & Controller &
                \rotatebox[origin=c]{90}{\textbf{NoNdiagonal M}} &
                \rotatebox[origin=c]{90}{\textbf{Uncertainty}} & 
                \rotatebox[origin=c]{90}{\parbox{2cm}{\centering\textbf{Constrained\\ variables}}} & 
                \rotatebox[origin=c]{90}{\parbox{2cm}{\centering\textbf{Unmeasured\\ states}}} & 
                \rotatebox[origin=c]{90}{\parbox{2cm}{\centering\textbf{Communication\\ bandwidth}}} & 
                \rotatebox[origin=c]{90}{\parbox{2cm}{\centering\textbf{Collision\\ avoidance}}} & 
                \rotatebox[origin=c]{90}{\textbf{Learning}} & 
                \rotatebox[origin=c]{90}{\textbf{Roll control}} &
                \rotatebox[origin=c]{90}{\textbf{Actuator faults}} & 
                \rotatebox[origin=c]{90}{\parbox{2cm}{\centering\textbf{Connectivity\\ preservation}}} & 
                \rotatebox[origin=c]{90}{\textbf{Input delay}}\\ 
                 \hline
                 \citep{BabiÄ‡2016361} & Adaptive & & & & & & $\times$ & & & & & \\ \hline
                 \citep{Bai20205015} & Sliding mode & & & & & & & & & & & \\ \hline
               \citep{bishop2012} & Redundant manipulator & $\times$ & & $\times$ & & & $\times$ & & & & & \\ \hline
                \citep{chen2019}  & Adaptive & & $\times$ & & & & & $\times$ & & & & \\ \hline
               \citep{Chen2015} & Backstepping & & & & & & $\times$ & & & & & \\ \hline
             \citep{chen2022_hierarchical} & Backstepping & $\times$ & & & & & & & & & & \\ \hline
             \citep{Chen2022}  & Adaptive & $\times$ & $\times$ & & & $\times$ & $\times$ & $\times$ & & & $\times$ & \\ \hline
                \citep{chen2020} & Sliding mode & & $\times$ & & & & &$\times$ & & & & \\ \hline
            \citep{Chen2015Global} & Backstepping & $\times$ & & & & & & & & & & \\ \hline
                \citep{dai2022}  &Barrier Lyapunov function and Backstepping  & $\times$ & & $\times$ & $\times$ & & $\times$& & & &$\times$ & \\ 
                 \hline
                \citep{do2016} & Backstepping & & $\times$ & & $\times$ & & & & & & & \\ \hline
               \citep{Do2012} &Backstepping& $\times$ & & & $\times$ & & $\times$ & & & & & \\ \hline
                 \citep{dong2021} & Backstepping & & $\times$ & & & & & & & & & \\ \hline
               \citep{Dong2010} & Lyapunov-based & & & & &$\times$ & & & & & & \\ \hline
              \citep{dong2022}  & Sliding mode& & $\times$& & $\times$ & & & & & & & \\ \hline
                 \citep{fahimi2007SlidingMode} &Sliding mode & & $\times$  & & & & & & & & & \\ \hline
               \citep{Fahimi2008} & Backstepping & & & & & & & & & & & \\ \hline
               \citep{Fahimi2007Nonlinear} & MPC & &$\times$  & & & & $\times$  & & & & & \\ \hline
                 \citep{Fan2017} & MPC& & &$\times$  & & & & & & & & \\ \hline
               \citep{Fu2022} & Finite-time-homogenity& & & & $\times$ & & & & & $\times$& & \\ \hline
               \citep{Fu2018Bioinspired} &Sliding mode & & & $\times$& & & & & & & & \\ \hline
                \citep{Fu2018Formation}  & Backstepping & & & & & & & & & & & \\ \hline
              \citep{Gao2023Fixed}  & Fixed-Time & & $\times$ & & $\times$ & $\times$ & & & & & & \\ \hline
              \citep{Gao2021}  & Backstepping & & $\times$& & $\times$ & $\times$ & & & & & & \\ \hline
              \citep{Ghommam2018}  & Barrier Lyapunov function and Backstepping & & $\times$& & $\times$ & & $\times$ & & & & $\times$ & \\ \hline
              \citep{Ghommam2020}  & Neural-networks & $\times$ & $\times$ & & $\times$ & & & $\times$ & & & & \\ \hline
              \citep{Ghommam2009}  & Backstepping & $\times$ &$\times$ & & & & & & & & & \\ \hline
              \citep{Gu2019Distributed}  & Backstepping & &$\times$ & & $\times$ & & &$\times$ & & & & \\ \hline
              \citep{Gu2022}  & Adaptive & &$\times$ & & & & & $\times$ & & & & \\ \hline
              \citep{Guo2021}  & Barrier Lyapunov function & $\times$ & $\times$ &  & $\times$ & & $\times$ & & & & $\times$ & \\ \hline
              \citep{Han2022}  & Adaptive & &$\times$ & & & & &$\times$ & & & & \\ \hline
              \citep{Hao2022}  & Backstepping & &$\times$ & & & & $\times$ & & & &$\times$ & \\ \hline
              \citep{He2021}  & Barrier Lyapunov function and Backstepping & $\times$& $\times$& & & & $\times$ &$\times$ & & &$\times$ & \\ 
                \hline
              \citep{He2022}  & Neural-networks  & $\times$&$\times$ & & & &$\times$ &$\times$ & & & $\times$ & \\ \hline
              \citep{Huang2021Adaptive}  &Adaptive & &$\times$ & &$\times$ & & &$\times$ & &$\times$ & & \\ \hline
              \citep{Huang2019Improved}  &Adaptive & &$\times$ & & $\times$ & & &$\times$ & & & & \\ \hline
              \citep{Huang2021Robust}  & Robust & &$\times$ & & & & & & & & & \\ \hline
              \citep{Huang2022}  & Backstepping& &$\times$ &$\times$ & $\times$ & & & & & & & \\ \hline
              \citep{Jiang2022NonfragileFormationControl}  & Lyapunov-based & $\times$ & $\times$ & & &$\times$ & & & & & & \\ \hline
              \citep{Jiang2022Sliding}  & Sliding mode &$\times$ & $\times$& & & & & & & & & \\ \hline
              \citep{Jiang2022Cooperative}  & Fuzzy & &$\times$ & & $\times$ & & &$\times$ & & & & \\ \hline
              \citep{Jiao2014}  & Sliding mode& & & & & & & & & & & \\ \hline
              \citep{Jin2022}  & Reinforcement learning & $\times$& & & & &$\times$ &$\times$ & & & & \\ \hline
              \citep{Jin2016}  & Barrier Lyapunov function& $\times$&$\times$ & & & & $\times$ & & & $\times$& $\times$ & \\ \hline
              \citep{kim2021}  & PID & & $\times$ & & & & & & & & & \\ \hline
              \citep{Li2019}  & Feedback linearization & &$\times$ & & & & & & & & & \\ \hline
              \citep{Li2018Adaptive}  & Adaptive& $\times$& & & & & & & & & & \\ \hline
              \citep{Li2018Finite}  & Sliding mode & & & & & & & & & & & \\ \hline
             \citep{Liang2019Novel}  & Robust & &$\times$ & & & &$\times$ & & & & & \\ \hline
              \citep{Liang2021SwarmVelocity}  & Barrier Lyapunov function & &$\times$ & & & &$\times$ & & & & & \\ \hline
              \citep{Liang2020Distributed}  & Neural-networks & &$\times$ & & & &$\times$ &$\times$ & & & & \\ \hline
               \citep{Liang2020Adaptive}  & Adaptive& &$\times$ & & & & $\times$ &$\times$ & & &$\times$ & \\ \hline
              \citep{Liang2022Secure}  & Feedback linearization & $\times$ & $\times$ & & & & & & & & & \\ \hline
              \citep{liang2019SwarmControl}  &Robust & &$\times$ &$\times$ & & & &$\times$ & & & & \\ \hline
               \citep{Lin2022Neural}  & Barrier Lyapunov function& $\times$&$\times$ & & $\times$ & &$\times$ &$\times$ & & &$\times$ & \\ \hline
              \citep{Liu2016Coordinated}  & Backstepping & &$\times$ & & & & &$\times$ & & & & \\ \hline
              \citep{Liu2021RobustEvent}  & Robust & &$\times$ & & & & &$\times$ & & & & \\ \hline
               \citep{Liu2017Saturated}  & Neural-networks& &$\times$ &$\times$ & $\times$ & & &$\times$ & & & & \\ \hline
              \citep{Liu2022RobustFuzzy}  & Fuzzy & &$\times$ & & & & &$\times$ & & & & \\ \hline
              \citep{Liu2022Distributed}  & MPC& & &$\times$ & & $\times$&$\times$ & & & &$\times$ & \\ \hline
              \citep{Liu2022Formation}  & Sliding mode& &$\times$ & & & & & & & & & \\ \hline
              \citep{Lu2018}  & Neural-networks & &$\times$ & & $\times$ & & &$\times$ & & & & \\ \hline
              \citep{Lv2021}  & MPC & &$\times$ &$\times$ & $\times$ & &$\times$ & & & & & \\ \hline
              \citep{Lv2023}  & MPC & &  &$\times$ & & &$\times$ &$\times$ & & & & \\ \hline
              \citep{Ma2019}  & Adaptive& & & & & & & $\times$& & & & \\ \hline
              \citep{Meng2012}  & Sliding mode& & & & & & & & & & & \\ 
               \bottomrule
	\end{tabular}}
\label{fig: limitation}
\end{table*}

\begin{table*}[H]
	\centering
	\scriptsize{	
        \begin{tabular}{{>{\centering\arraybackslash}m{2.165em}
                >{\centering\arraybackslash}m{22.15em}
                ccccccccccc}}\toprule
        \multicolumn{13}{c}{\textbf{Simulation}} \\ \hline
                Ref    & Controller &
                \rotatebox[origin=c]{90}{\textbf{NoNdiagonal M}} &
                \rotatebox[origin=c]{90}{\textbf{Uncertainty}} & 
                \rotatebox[origin=c]{90}{\parbox{2cm}{\centering\textbf{Constrained\\ variables}}} & 
                \rotatebox[origin=c]{90}{\parbox{2cm}{\centering\textbf{Unmeasured\\ states}}} & 
                \rotatebox[origin=c]{90}{\parbox{2cm}{\centering\textbf{Communication\\ bandwidth}}} & 
                \rotatebox[origin=c]{90}{\parbox{2cm}{\centering\textbf{Collision\\ avoidance}}} & 
                \rotatebox[origin=c]{90}{\textbf{Learning}} & 
                \rotatebox[origin=c]{90}{\textbf{Roll control}} &
                \rotatebox[origin=c]{90}{\textbf{Actuator faults}} & 
                \rotatebox[origin=c]{90}{\parbox{2cm}{\centering\textbf{Connectivity\\ preservation}}} & 
                \rotatebox[origin=c]{90}{\textbf{Input delay}}\\ 
                \hline
                \citep{Mu2020}  & Adaptive& &$\times$ &$\times$ & &$\times$ & &$\times$ & & & & \\ \hline
                \citep{Ning2022}  & Adaptive& &$\times$ & & $\times$ & & &$\times$ & & & & \\ \hline
               \citep{Panagou2013}  & Backstepping & & & & & &$\times$ & & & &$\times$ & \\ \hline
              \citep{Park2022}  &Neural-networks &$\times$ &$\times$ & & $\times$&$\times$ & & $\times$& & & & \\ \hline
              \citep{Park2021}  & Backstepping &$\times$ &$\times$ & & & &$\times$ & & & &$\times$ & \\ \hline
              \citep{Peng2012NeuralAdaptive}  & Adaptive & &$\times$ & & & &$\times$ &$\times$ & & & & \\ \hline
              \citep{Peng2011RobustAdaptive}  &Adaptive &$\times$ &$\times$ & &$\times$ & & &$\times$ & & & & \\ \hline
              \citep{Peng2013AdaptiveDynamic}  & Adaptive&$\times$ &$\times$ & & & & &$\times$ & & & & \\ \hline
               \citep{Peng2012RobustLeader}  & Backstepping & $\times$&$\times$ & & & & &$\times$ & & & & \\ \hline
              \citep{Peng2019PathGuided}  & Adaptive & &$\times$ & & & &$\times$ &$\times$ & & & &\\ \hline
              \citep{Qin2017}  & Sliding mode & &$\times$ & & & &$\times$ & & & & & \\ \hline
               \citep{Riahifard2020}  & Adaptive& &$\times$ &$\times$ & & & & & & & & \\\hline
              \citep{Shojaei2015}  & Adaptive & &$\times$ &$\times$ & & & &$\times$ & & & & \\ \hline
              \citep{Song2022}  & Sliding mode & &$\times$ &$\times$ & & $\times$ & &$\times$ & & & & \\ \hline
              \citep{Sun2020}  & MPC & & &$\times$ & & &$\times$ & & & & & \\ \hline
              \citep{Sun2018}  & Sliding mode& &$\times$ & & & & & & & & & \\ \hline
               \citep{Sun2022}  & Sliding mode & &$\times$ & &$\times$ & &$\times$ &$\times$ & & & & \\ \hline
              \citep{Wang2022}  & Robust& $\times$& & & & &$\times$ & & & & & \\ \hline
              \citep{Wang2019}  & Barrier Lyapunov function& &$\times$ &$\times$ & & &$\times$ &$\times$ & & & & \\ \hline
               \citep{Wang2021}  & Barrier Lyapunov function& &$\times$ & & $\times$ &$\times$ &$\times$ & & & & $\times$ & \\ \hline
              \citep{Wang2012}  & Adaptive& $\times$&$\times$ & & & & &$\times$ & & & & \\ \hline
                 \citep{Wu2022}  &Barrier Lyapunov function& &$\times$ &$\times$ & $\times$ & &$\times$ &$\times$ & & & $\times$ & \\ \hline
              \citep{Wu2021}  & Adaptive& &$\times$ & & $\times$ & & &$\times$ & & & & \\ \hline
              \citep{Wu2023}  & Barrier Lyapunov function& &$\times$ &$\times$ & $\times$ & &$\times$ &$\times$ & & & $\times$ & \\ \hline
              \citep{Xia2021}  & Backstepping & &$\times$ &$\times$ &$\times$ & &$\times$ &$\times$ & & & & \\ \hline
               \citep{Xia2022}  & Backstepping & &$\times$ &$\times$ &$\times$ & &$\times$ & & & & & \\ \hline
              \citep{Xia2020}  & Barrier Lyapunov function& &$\times$ &$\times$ &$\times$ & & $\times$ & & & &$\times$ & \\ \hline
              \citep{Xiong2022}  & Barrier Lyapunov function& $\times$&$\times$ &$\times$ & & &$\times$ &$\times$ & & & $\times$& \\ \hline
              \citep{Ye2021}  & Finite-time-homogeneity & & & & & & & & & & & \\ \hline
              \citep{Yu2019}  & Lyapunov-based & & $\times$ & & $\times$ & & & & & & & \\ \hline
              \citep{Yue2019}  & Robust & & & & &$\times$ & & & &$\times$ & & \\ \hline
               \citep{Yuhan2022}  & Adaptive & &$\times$ &$\times$ & & & &$\times$ & & & & \\ \hline
               \citep{Zhang2022EventTriggered}  & Adaptive & &$\times$ & & &$\times$ & &$\times$ & & & & \\ \hline
              \citep{Zhang2020FormationControlMultiple}  & Backstepping & & & & & & & & & & & \\ \hline
              \citep{Zhang2021LVS}  & Adaptive& &$\times$ & &$\times$ &$\times$ & &$\times$ & &$\times$ & & \\ \hline
              \citep{Zhang2022Bearing}  & Backstepping & &$\times$ & & & & &$\times$ & & & & \\ \hline
               \citep{Zhang2021NovelEventTriggered}  & Adaptive & &$\times$ & & &$\times$ & &$\times$ & & & & \\ \hline
              \citep{Zhang2022AdaptiveDistributed}  & Adaptive & & $\times$ & & &$\times$ & &$\times$ & & & & \\ \hline
              \citep{Zhang2022RobustAdaptive}   &Adaptive & &$\times$ & & $\times$ & & & $\times$& & & & \\ \hline
               \citep{Zhao2021}  & Neural-networks& $\times$& & & & & &$\times$ & $\times$ & & & \\ \hline
              \citep{Zhou2022}  & Adaptive & & $\times$ & & $\times$ &$\times$ & &$\times$ & & & & \\ \hline
              \citep{Zhuang2019}  & Sliding mode & & & & & &$\times$ & & & & & \\  \bottomrule
\multicolumn{13}{c}{\textbf{Experiment}} \\ \hline
 \citep{Arrichiello201273} & Null-Space-based Behavioral & & & & & & $\times$ & & & & & \\ \hline
\citep{BÃrhaug2011493} & Feedback linearization & $\times$ & & $\times$ & & & & & & & & \\ \hline
\citep{Chen2022reinforcement} & Sliding mode & & & $\times$ & & & $\times$ & $\times$ & $\times$ & $\times$ & & $\times$\\ \hline
\citep{Eek2021171} & Sliding mode & $\times$ & & & & & $\times$ & & & & & \\ \hline
\citep{Gu2019Antidisturbance} & Backstepping & & $\times$ & & $\times$ & & & & & & & \\ \hline
\citep{Hu2022} & PID & & $\times$ & & $\times$ & & & & & & & \\ \hline
\citep{hu2022bearing} & Backstepping & &  &$\times$ & & & & & & & & \\ \hline
\citep{Hu2021} & Adaptive & & $\times$ & & $\times$ & & & & & & & \\ \hline
\citep{hu2023spontaneous} & Backstepping & & &  & &  & & & & & & \\ \hline
\citep{Li2020} & Feedback linearization & & $\times$ & & & & & & & & & \\ \hline
\citep{Liu2020Collective} & PID & & & & & & & & & & & \\ \hline
\citep{Liu2022Scanning} & MPC & & & $\times$ & & & & & & & & \\ \hline
\citep{Lv2022} & Backstepping & & $\times$ &  & $\times$ & $\times$ & & & & & & \\ \hline
\citep{Ringback2021} & PID & & & & & & $\times$ & & & & & \\ \hline
\citep{Tang2023} & PID & & & & & & & & & & & \\ \hline
\citep{Zhang2019FormationControlAutonomous} & Fuzzy& & & & & & & & & & & \\ \hline
\citep{Zhou2020} & Adaptive & & $\times$ & $\times$ & & & & $\times$ & & & & \\
\bottomrule
	\end{tabular}}
\label{fig: limitation2}
\end{table*}

\subsubsection{Unmeasured states} \label{subsubsec:Unmeasured}

One of the initial steps in designing any controller is to understand the available information during each time instant, along with its reliability. Although in numerical validation this may not be a decisive factor, if one intends to bring a controller to experimental validation, it becomes a vital consideration. Despite the continuous evolution of sensor technology, challenges still exist in accurately measuring velocity or disturbances in real-time. This has led to the proposition of approaches where from the design stage, it is assumed that some information is not available and needs to be reconstructed. For this purpose, two main approaches have been identified: the use of state observers or machine learning techniques. However, in this section, we will focus on the former branch, leaving the latter for Section \ref{subsec:Learning}.

Out of the 125 articles, 32 take into account unmeasurable states and implement a different approach than machine learning to address them. Among them, 21 articles utilize extended state observers with various variations, 6 employ Disturbance-Observer-Based methods, and finally, 5 implement High-Gain Observer techniques. On the other hand, among the variables that are most commonly reconstructed are various external disturbances, own or other agents' velocities, positions and orientations of neighbors or targets, or the unknown sideslip. Below, there is a summary of each of these approaches and how they contribute to the coordinated control system.

Disturbances are one of the most challenging variables to measure and have a direct impact on vessel behavior. The use of Disturbance-Observers allows estimating these values from position or velocity measurements and also compensating them in control. Some examples of this can be found in \citep{Do2012}, \citep{do2016}, and \citep{Wu2021}, where the estimated values are included in control to compensate for disturbances such as wind, currents, and sea loads. On the other hand, \citep{dong2022} implements a similar nonlinear observer, but this time it works in conjunction with a sliding-mode controller. Finally, an innovative approach is presented in \citep{Huang2019Improved}, where Minimal Learning Parameters are included to enhance the observer's performance, thus proposing an adaptive finite-time disturbance observer.

Similarly, high-gain observers are primarily used for the estimation or reconstruction of velocities, which in many cases are difficult to measure accurately. For instance, \citep{Wang2021} and \citep{Lin2022Neural} implement this type of observer in followers aiming to estimate or reconstruct the velocities of leaders, integrating this information with tan-type  BLF controllers. This is akin to \citep{Ghommam2018}, where a fast convergent observer was designed. Conversely, \citep{Lu2018} implements the observer in the leader to estimate the follower velocities, and finally, \citep{Huang2022} implements it locally to estimate the own velocity and achieve feedback.

To conclude, extended state observers are the most commonly used, primarily due to their robustness, to the extent that some authors implement more than one within their control schemes. This is evident in cases like \citep{Gao2023Fixed} and \citep{Gao2021}, where two observers are included, the first focused on disturbances and uncertainties, and the second focused on reconstructing the velocities and positions of the target. Another approach is the implementation of reduced-order extended state observers, which in \citep{Liu2017Saturated} was used to estimate the unknown sideslip and thus compensate for the undesired effect of increasing sway. Regarding control techniques, it is evident that observers have been integrated with various control techniques. For example, \citep{Lv2021} designs an observer to calculate disturbances and integrates it with MPC, while \citep{Wu2023} implements the observer to estimate velocities and integrates it with BLF.

\subsubsection{Collision avoidance and Connectivity preservation} \label{subsubsec:Collision}

In this subsection, we will address jointly two problems that, initially, may appear to lack a direct connection. However, after the second phase of the present review, a clear relationship between them was evidenced due to the mathematical approach to solving them. In this approach, collision avoidance is addressed as a minimum or safe distance between agents, while connectivity preservation is defined as a maximum distance, with both constraints remaining constant over time. However, there is a set of articles that do not address them simultaneously. Therefore, the aim is to highlight both the techniques or approaches that address them jointly  and those that treat them independently, particularly collision avoidance.
Out of the 125 articles, 19 address both problems simultaneously, and there is a clear recurrence in the use of the Barrier Lyapunov Function (BLF) technique within this group, with a total of 12 articles implementing it. However, some details vary among different authors. For example,  \citep{Liang2020Adaptive},  \citep{Jin2016}, \citep{He2021}, \citep{Wang2021}, and  \citep{Lin2022Neural} incorporate tanh type BLF to handle constraint requirements within the range generated by a LOS scheme, contributing to achieving the desired formation. Another similar case is  \citep{Wang2019}, who implements two tanh-type BLFs to keep the formation error within a predefined range. On the other hand,  \citep{Guo2021} and  \citep{dai2022} also use LOS to generate constraints but implement Asymmetric Barrier Lyapunov Functions to handle them.

Although the LOS scheme is widely used in various articles, other interesting approaches found in the literature that are also capable of satisfying both constraints are: \citep{Wu2022} formulates constrained quadratic optimization problems using Lyapunov functions, high-order control barrier functions, and stable reduced-error observers to generate control actions without violating input constraints, stability, and safety. On the other hand, in \citep{Xiong2022}, a strategy is implemented to handle collision and connectivity constraints by transforming the error into a dynamic surface through the creation of a dynamic surface that ensures stability with Lyapunov barrier functions.

Among the articles addressing these constraints simultaneously, the control schemes proposed in \citep{Liu2022Distributed}, \citep{Chen2022}, and \citep{He2022} stand out. \citep{Liu2022Distributed} formulates a Nonlinear Model Predictive Control (NMPC) with connectivity preservation and collision avoidance as a quadratic optimization problem with input and output constraints, while \citep{Chen2022} and  \citep{He2022}  employ the Prescribed Performance Control (PPC) methodology to ensure both constraints. It is important to note that these articles are the only ones that do not use Lyapunov barrier functions to tackle these constraints. Additionally, the constraints proposed in \citep{Panagou2013} and \citep{Jin2016} are noteworthy, where the main variation is the addition of a field-of-view constraint. This is because the leader's information is not transmitted directly from the leader but measured by local sensors in the follower.

On the other hand, out of the 125 articles, 25 address collision avoidance without considering connectivity preservation. There is a clear recurrence of the artificial potential functions technique within this group to address this problem, with a total of 16 articles implementing it. The main difference among this group of articles lies in the technique they are combined with to complete the control scheme. Some examples include \citep{Peng2012NeuralAdaptive} and \citep{Peng2019PathGuided}, where adaptive controls are used, or \citep{Do2012} and \citep{Xia2021}, where they are combined with backstepping, or \citep{Liang2019Novel}, \citep{Sun2022}, \citep{Qin2017}, and \citep{Zhuang2019}, where they are combined with sliding mode techniques.

Continuing the analysis of the remaining 9 articles, the use of techniques such as Null-Space-based Behavioral (NSB) in \citep{Arrichiello201273} and \citep{Eek2021171} is highlighted. In addition to including collision avoidance, other tasks are defined, such as ensuring distance between vehicles. Although this approach is less common, it can be adapted to more complex tasks with different objectives and priorities. Additionally, Model Predictive Control is employed in \citep{Fahimi2007Nonlinear}, \citep{Lv2021}, and \citep{Lv2023}, where distance constraints are included. Reinforcement Learning is also utilized in \citep{Jin2022} and \citep{Chen2022reinforcement}. Finally, an innovative approach compared to the previously mentioned ones is implemented in \citep{Wang2022}, which is based on elliptic boundary cycles.

\subsubsection{Communication considerations} \label{subsubsec:Communication_considerations}

In most articles, the number of vehicles and the information shared among them is not a limitation, primarily because their validations are numerical simulations. However, a first step towards reality is to propose communication topologies where information is not shared with all agents but only with some neighbors or the leader. Broadcast communication, although used in fleet control, can lead to problems such as information redundancy, causing network nodes to receive redundant and non-useful information, not to mention the energy cost of generating it. On the other hand, communication topologies with neighbors or through graphs ensure that each node receives essential information to calculate its control actions based on the information from its closest neighbors, aiming to avoid collisions and preserve formation or safe distances between vehicles. Similarly, communication with neighbors or the leader reduces interference and can be more scalable as the number of vessels in a fleet increases.

Nevertheless, although many of the reviewed articles implement communication among neighbors or with the leader, as mentioned in Section \ref{subsec:RQ3}, this section will address other considerations related to communications. These include approaches such as reducing shared information, asynchronous communications, communication delays, bandwidth reduction towards actuators, and real-time optimization of topologies.

Among the articles considering bandwidth, the commonly used approach is asynchronous transmission. This mechanism dynamically adjusts the frequency of data transmission or reception towards neighbors or leaders based on the change or variation of some variable. Its main objective is to improve efficiency and management in information exchange without compromising the performance of coordinated control. For example, \citep{Zhang2022AdaptiveDistributed} and \citep{Zhou2022} implement asynchronous communications where they update the position and angle of each agent based on the change in velocity, while \citep{Chen2022} updates values based on the variation in control action between each instant. Another case is presented in \citep{Gao2023Fixed} and \citep{Gao2021}, where state observers are used to estimate the position and velocity of the target, and event triggers are designed to update information based on the error. On the other hand, \citep{Lv2022} and \citep{Song2022} perform updates based on a local path variable predictor and the internal dynamic auxiliary variable, respectively.

Another less direct approach is reducing transmitted information through state estimation. For example, \citep{Mu2020} implements a virtual leader through which it estimates the velocity of the real leader based on its position and orientation. Similarly, \citep{Wang2021}, \citep{Park2022}, and \citep{Zhang2021LVS} propose the use of state observers in followers to estimate the velocities and positions of the leaders. It is important to highlight that the latter two utilize neural networks for their estimation. On the other hand, \citep{Dong2010}, \citep{Yue2019} and \citep{Jiang2022NonfragileFormationControl} assess the robustness of their controllers against communication delays and packet losses. Finally, an innovative approach is presented in \citep{Liu2022Distributed}, where the Floyd method is included to optimize the topology in order to achieve optimal online communication.

To conclude with communication optimization, another approach identified in the literature is the reduction or optimization of transmissions to actuators or updates in control actions. Although this differs from communication between fleet agents, this strategy may be of interest in other applications where computational capacity is a limiting factor or where periodic variation in control actions is undesirable. An example of this is implemented in \citep{chen2020}, where control actions are enabled based on their change from the previous instant. This is very similar to the proposal by \citep{Lv2022}, which additionally uses Event-Triggered mechanisms for changes in actuators and state estimation.

\subsubsection{Other considerations} \label{subsubsec:Other}

Finally, we will address some less common but significant considerations that are worth mentioning due to their impact on framing future work. For instance, the proposition of control laws with 4 degrees of freedom including Roll is an example. While most authors simplify to 3 degrees of freedom for planar motion control, it cannot be denied that the rolling angle and its velocity can become destabilizing factors under certain disturbances, particularly in vessels equipped with a rudder to control yaw. For this reason, \citep{Chen2022reinforcement} proposes mitigating the effects of roll motion by employing a rudder stabilization strategy through the implementation of a finite-time Sliding Mode Control (SMC) scheme based on Reinforcement Learning (RL). In this same approach, a consideration that few authors have addressed is also included, such as the actuator delay, which can affect the performance of the controllers. To counteract this, the Pade approximation strategy and Laplace transform are used, defining an auxiliary variable to compensate for the actuator delay.


\subsection{RQ.2: Which are the most important characteristics about techniques that have been experimentally tested?} \label{subsec:rq2}

The aim of this section is to gain insights into the progress and state of the art of proposals that have been applied and validated in real-world settings and diverse contexts, and therefore this section focus on those papers that underwent experimental validation. This issue becomes particularly relevant when verifying the number of papers that have been experimentally validated. Out of the 125 final articles, only 17 present experimental results. This demonstrates a clear research gap, which can be attributed at first glance to the challenges of implementing control algorithms on real autonomous platforms, due to complexity and associated costs and resources.

Nonetheless, in recent years, there has been a clear increase in the design of various new models of ASVs or Unmanned Surface Vehicle (USV)s, as evidenced in  \citep{zhang2021survey}. This suggests that other factors may hinder the implementation and validation of different control algorithms. Based on the discussion in Section \ref{subsec:rq1.1}, it can be concluded that many of the assumptions made during the design phases of the controllers are significant. This implies that numerical validation in a controlled environment, where the limitations and disturbances of the system are well defined, may be feasible. On the other hand, experimental validation, where various external factors exist, is more complex and presents different challenges that need to be considered in both pre- and post-validation phases.

Upon examining Table 8 in the experiments section, it is evident that the most recurring assumptions in the literature are uncertainty and variable constraints, with 6 and 7 articles, respectively. It is imperative to consider these assumptions in a control system that will face uncontrollable external conditions or factors and also presents physical limitations. Furthermore, it is relevant to highlight assumptions such as non-diagonal M, communication bandwidth, or connectivity preservation, which also impact controller performance and can be mitigated with a symmetrical design in ASV models to reduce uncertainty in non-diagonality or by limiting the number of vessels to avoid bandwidth saturation during experiments.

Similarly, actuator faults or input delays are the least frequent assumptions within the experimental articles. For both cases, these assumptions may be easily negligible or controllable during short-duration experiments. However, they gain importance when considering robust systems capable of performing long-duration tasks or operating in adverse environments. A similar situation arises with disturbances caused by wind, waves, or currents, which are not directly controllable during outdoor experiments. Nonetheless, it is feasible to mitigate their impact by selecting time intervals when environmental conditions are favorable and minimally affect controller performance.

Continuing with the analysis of experimental validation and its applications, Table \ref{fig: data_experimental} presents relevant information from the 17 articles. This includes the proposed control approach, the application, the environment or experimental setting, reference trajectories or paths, a brief description, and the number of vessels used during the experiment. Although these validations, in general, are not framed as tests where vessels must perform a specific task, it is worth noting the case of  \citep{Arrichiello201273}, where the main objective is to capture an object using a rope connected to two boats and transport it to a specific sector.

When analyzing the various control approaches in the experimental articles, it proves challenging to identify clear recurrences or trends. However, techniques such as sliding mode, Lyapunov, and adaptive control have been widely applied. In contrast, nonlinear approaches or MPC are less common in this type of validation. Nonetheless, the presence of PID or fuzzy controls is notable. Despite being simpler and less robust than other techniques, they have been successfully implemented in real tests, where faster response times and processing are required. Another commonly used technique in experimental validation is extended state observers, which are often utilized to estimate unmeasurable velocities or disturbances. This proves very useful in platforms with limited sensors, as seen in \citep{Gu2019Antidisturbance}, or for optimization in communications, as demonstrated in \citep{Lv2022} and \citep{Liu2020Collective}.

\begin{table*}[H]
\caption{Experimental references: control approaches and vessel configurations}
	\centering
	\scriptsize{	
        \begin{tabular}{{>{\centering\arraybackslash}m{2.165em}
                >{\centering\arraybackslash}m{10.15em}
                >{\centering\arraybackslash}m{4.50em}
                >{\centering\arraybackslash}m{2.15em}
                >{\centering\arraybackslash}m{4.15em}
                >{\centering\arraybackslash}m{26.15em}
                >{\centering\arraybackslash}m{2.50em}}}\toprule
                   \textbf{Ref} & \textbf{Approach} 
                   & \textbf{Application}  & \textbf{Test place} 
                   & \textbf{Trajectory} & \textbf{Description} 
                   & \textbf{\# ASVs}\\ \midrule
        \citep{Arrichiello201273}   & A multi-layered control architecture
        & Collection    & Lake      & Straight-line           
        & Two under-actuated ASVs are employed to capture a target and transport it to a specified goal. These vehicles are interconnected by a floating flexible   rope.& 2      \\

         & & & & & & \\
        
         \citep{BÃrhaug2011493}       & A nonlinear synchronization control law.    & Triangular formation        & Pool                      & Straight-line           & Three vessels were allocated to distinct straight-line paths (one real and two virtual), aligned longitudinally in the pool, and instructed to maintain an inline formation. & 1                \\

          & & & & & & \\
        
        \citep{Chen2022reinforcement}       & Sliding mode control scheme based on reinforcement learning  & Leader-follower & Lake 
        & Straight-line and arc & Two   USVs are designated to achieve a close formation, with one acting as the leader and the other as the follower. The formation trajectory consists of   both straight-line segments and arcs. & 2  \\

        & & & & & & \\
        
        \citep{Eek2021171}       & Adaptive feedback linearizing PD-controller with sliding-mode & Parallel formation & Sea                    & Straight-line and curv      & A   formation control method is implemented for two USVs to track curved paths while navigating in the presence of ocean currents. The experiments were   conducted under calm sea state conditions. & 2                \\

        & & & & & & \\

        \citep{Gu2019Antidisturbance}      & Anti disturbance kinetic control laws are a dynamic surface   control & Parallel   formation & Lake      & Straight-line and arc   & Three   ASVs are allocated to three parallel paths, and they achieve synchronized   formation. The trajectories consist of straight segments with semicircular turns.& 3                \\

        & & & & & & \\

        \citep{Hu2022}         & A distributed estimator for each vessel                                        & Surrounding   control         & Lake                      & Circle                        & Three   vessels capture and rotate evenly around a moving target located at the   center, operating on an intermittent communication network. & 3                \\

        & & & & & & \\

        \citep{Hu2021}          & Adaptive distributed control law is designed for the MASs                      & Surrounding   control         & Pool                      & Square   Formation            & The   group comprises six vessels with a sensing radius R set at 600mm and a safe   radius r set at 200mm. The first four vessels act as followers, while the remaining two serve as targets.    & 6                \\

        & & & & & & \\

        \citep{hu2023spontaneous}          & Distributed guiding-vector-field algorithm                              & Leader-followers              & Pool                      & Euclidean curves            & This   paper presents a Distributed Guiding Vector-Field algorithm designed for a team of ASVs. The algorithm enables the formation of a   spontaneous-ordering platoon, guiding their movement along a predefined   desired path within n-dimensional Euclidean space & 3  \\

        & & & & & & \\

        \citep{hu2022bearing}         & A bearing-only estimation term to approximate the target state                 & Surrounding control         & Pool                      & Circle                        & During the experiments, a multi-USV system comprising three USVs and a target vessel is utilized. The USVs start from initial positions with varying orientations and aim to surround a fourth USV designated as the target. & 4  \\

        & & & & & & \\

        \citep{Li2020}         & Cascade system theory and Lyapunov stability and Adaptive LOS                  & Triangle formation          & Sea                       & Straight-line                 & The expected relative distance and direction among USVs are 300 meters and an angle of 135 degrees with respect to the central axis, with a triangular formation along a straight-line trajectory. & 3                \\

        & & & & & & \\

        \citep{Liu2020Collective}         & $\tau_{1}$ is (PI) form with proper feedforward compensation and $\tau_{2}$   is of the PD & Surrounding   control         & Lake                      & Concentric   circles          & USVs   chased and surrounded a target vessel within 50 seconds on a 40 × 40 m2 water surface. They maintained fixed distances and positions around the target   until the mission was completed by the 160th second. & 3           \\

        & & & & & & \\

        \citep{Liu2022Scanning}         & Lyapunov-based model predictive control path-following                  & Channel   boundaries          & Lake                      & Straight-line   and curv      & Two   leaders and one follower start randomly on a 15-meter radius water area.   Leaders follow planned paths along water channel boundaries, while the   follower tracks the estimated path trajectory with local communication. & 3 \\

        & & & & & & \\

        \citep{Lv2022}         & Based on event-triggered ESO  the dynamic positioning                   & Parallel   formation          & Lake                      & Concentric   circles          & Three   ASVs follow parallel trajectories on the lake, forming concentric circles.   The vehicles are equipped with  global positioning system (GPS) with a precision of 2 meters.                                                                                                                           & 3                \\

        & & & & & & \\
        \citep{Ringback2021}         & Potential function                                                             & Surrounding   control         & Lake                      & Concentric   triangle         & Three   ASVs seek to reach a target and surround it in an equilateral triangle   formation, with the target remaining in the center. & 3                \\

        & & & & & & \\
        
        \citep{Tang2023}         & Formation Tracking Control Protocol and Curvature Observer                     & Channel   boundaries          & Pool                      & Straight-line   and curv      & A   group of three ASVs navigates through a narrow, curved canal, with   theoretically defined boundaries using the Flexible Formation Tracking   Control Protocol (FFTC).  & 3\\

        & & & & & & \\
        
        \citep{Zhang2019FormationControlAutonomous}         & Fuzzy PID controller for ASVs                                                  & Formation   diamond           & Lake                      & Straight-line   and circle    & The   experiments with 4 ASVs involve tracking a straight trajectory in diamond   formations and then switching to an in-line formation.       & 4                \\

        & & & & & & \\

        \citep{Zhou2020}         & Adaptive fuzzy method and lyapunov                                             & Leader-follower,   triangular & Lake                      & Straight-line   and waypoints & Three   USVs participate, with one serving as the leader and the remaining two as   followers. The desired path comprises multiple points, and the leader vehicle   maintains a desired forward speed of 1.5 m/s. & 3     \\ 
        
        \bottomrule
	\end{tabular}}
\label{fig: data_experimental}
\end{table*}

\begin{table*}[H]
\caption{Experimental references: technical description of the vehicles}
	\centering
        \scriptsize{
		\begin{tabular}{{>{\centering\arraybackslash}m{6.165em}
                >{\centering\arraybackslash}m{2.15em}
                >{\centering\arraybackslash}m{4.50em}
                >{\centering\arraybackslash}m{2.15em}
                >{\centering\arraybackslash}m{4.15em}
                >{\centering\arraybackslash}m{6.15em}
                >{\centering\arraybackslash}m{26.15em}}}\toprule
                   \textbf{Model} & \textbf{Ref} & \textbf{Size (len/wid)}   & \textbf{Weighs} & \textbf{Actuators}  & \textbf{Communications}               & \textbf{Description}  \\ \midrule                                                                                   
                OceanScience QBoat-I  & \citep{Arrichiello201273}            & \begin{tabular}[c]{@{}c@{}}2,13   m /\\      0,71 m\end{tabular}  & 48,0   Kg       & Thrusters   and a rudder & 47M   Hz RC link                      & The   ASVs feature a u-blox EVK-5H GPS for global position updates at 2Hz and a   Microstrain 3DM-G IMU with an integrated compass sampled at 50 Hz. Control   software is developed using the open-source Robot Operating System. \\

                 & & & & & & \\
                 
                CyberShip II CS2      & \citep{BÃrhaug2011493}& 1,255   m / 0,29 m  & 23,8   Kg       & Thrusters                & N/S   & CyberShipII,   a fully-actuated supply vessel, is outfitted with two rudders, two aft   propellers, and a bow thruster. During experiments, only the aft propellers   and bow thruster were utilized. Thrusters are controlled by DC motor   controllers with realistic response times and saturation levels tailored to   the vessel's size and weight.\\
                
                 & & & & & & \\
                 
                N/S & \citep{Chen2022reinforcement} & N/S  & N/S             & Thrusters                & Wireless   AP station           & Onboard   sensors comprise a compass/GPS receiver, laser radar, and CCD camera. The   compass/GPS receiver provides the position and attitude of the USV. Laser   radar is utilized for detection and information acquisition \\

                 & & & & & & \\
                 
                Odin and Frigg & \citep{Eek2021171} & N/S  & N/S             & Waterjet   system        & N/S                             & Odin,   a small vessel tailored for near-shore operations, must autonomously execute   all missions in close proximity to the shore. Navigating near the shore poses   significant safety challenges, necessitating a route planner that ensures   safety without compromising operational efficiency. \\
 
                & & & & & & \\
                
                HUSTER-12   &     \citep{Liu2020Collective} \citep{Liu2022Scanning}  \citep{Hu2022} & \begin{tabular}[c]{@{}c@{}}1,60   m /\\       0,45 m\end{tabular} & 25   Kg         & Thrusters   and a rudder  & 433   M wireless sensor (E32-433T30S) & Each   vessel is outfitted with an onboard differential GPS receiver   (NovAtel-OEM615), a differential GPS wireless sensor (E32-400T20S), two differential GPS antennas, a   433 MHz wireless sensor (E32-433T30S), an accelerometer-gyroscope (AG) chip   (MPU6050), an embedded controller (STM32F407VGT6), a motor driver   (SEAKING-V3), a high-power battery (RUIYI-48), and a waterjet propeller   (HJ-064).                              \\

                & & & & & & \\
                 
                CSICET-DH01 & \citep{Lv2022}  \citep{Gu2019Antidisturbance} & 
                \begin{tabular}[c]{@{}c@{}}1,10   m /\\       0,36 m\end{tabular} & 5,4   Kg        & Thrusters   and a rudder  & ZigBee network                        & Each   ASV is furnished with a global navigation satellite system receiver,   an attitude sensor, a motion coprocessor, a microcontrol unit, and a   wireless communication device. Position data is acquired via a GNSS receiver   operating at a 10-Hz sample rate. \\

                DUCKLINGS             &  \citep{Ringback2021}
                & \begin{tabular}[c]{@{}c@{}}1,12   m /\\       0,72 m\end{tabular} & 15,0   Kg       & Thrusters                & Radio   Airmax 5GHz 8dBi              & DUCKLINGS   are compact catamarans propelled by two external thrusters located at the   rear. Their small size enables agile steering and rapid initial acceleration.   The microcontrol unit oversees navigation and path-following using GPS and an inertial   measurement unit, and responds to a predefined set of commands. \\

                 & & & & & & \\
                 
                N/S                   & \citep{Zhang2019FormationControlAutonomous}                                         & N/S   & N/S             & Thrusters   and a rudder  & ZigBee network DRF1605H               & Each   ASV utilizes an STM32 microcontroller unit as the computing core, responsible   for executing the control algorithm. The navigation system relies on a GPS   module and its antenna to obtain real-time position data for the ASVs. For   communication, a DRF1605H module, a type of ZigBee module, is employed.\\

                 & & & & & & \\
                 
                N/S              &  \citep{Zhou2020}    & N/S                                                               & N/S             & Thrusters                & wireless   network                    & The   platform consists of a carbon fiber hull housing the electronic system,   actuators, power, and communication equipment. The actuators, located at the   stern of the USV, control its yaw through differential action. The onboard   system includes the control system, Inertial Navigation System (INS), and GPS.   Power is supplied by lithium batteries, and communication is conducted   through a wireless network. \\

                 & & & & & & \\
                
                HUSTER-0.3            & \citep{Hu2021}    \citep{hu2023spontaneous}  \citep{hu2022bearing}    & \begin{tabular}[c]{@{}c@{}}0,3   m / \\      0,12 m\end{tabular}  & N/S             & Thrusters                & 2,4GHz   wireless network             & Each   USV is outfitted with two 5V DC motors, two transmission shafts measuring   150mm each, two speed encoders (Freescale Mini256ABC), onboard 2.4GHz   wireless components (NRF24L01), three infrared emitters with a peak   wavelength of 850nm, two LED indicators, and an LI-PO Battery (7.4V, 1300mAh,   25c). These components are all controlled by a microcontroller   (STM32F103CBT6).                              \\

                 & & & & & & \\

                Jellyfish             & \citep{Li2020}   & N/S  & N/S & N/S                      & N/S  & Jellyfish  USVs developed by the research group were used in a USV formation cruising   control test in a sea area adjacent to Guzhenkou Bay in Qingtao.  \\

        \bottomrule
    \end{tabular}}
\label{fig: data_model_experimental}
\end{table*}

Regarding the applications, they have been identified and grouped into four main categories: leader-follower, formation (triangular, diamond, parallel), surrounding control, and channel boundaries, with formation being the most recurrent. It is important to note that the main difference between the channel boundaries applications and the others lies in their consideration of navigation limits, aiming to ensure that the formation adapts without colliding with them. Although coordinated leader-follower techniques, as described in \citep{Liu2022Scanning}, or formation tracking, as mentioned in \citep{Tang2023}, are proposed to achieve this goal, among all the previously mentioned classifications, the latter is one that takes into account more realistic factors, such as fleet mobilization through port channels or narrow streets.

In contrast to simulation validation, where disturbances are defined and known, experimental validation primarily depends on the test site and the weather conditions at the time of evaluation. Upon examining various locations, it is evident that lakes are preferred for this type of validation, primarily due to the presence of disturbances such as waves, wind, and currents, which, compared to other environments, are of lesser magnitude. On the other hand, the two additional options represent opposing extremes: validation in indoor pools, where external disturbances are virtually negligible, and testing in open seas, where disturbances are considerably more aggressive and require more robust equipment to withstand them.

Another important aspect of the existing experimental validations is the number of vehicles composing the fleet and the reference trajectories or paths to follow. Concerning trajectories, the article presenting the most complex ones compared to others has been \citep{hu2023spontaneous}, which proposes tracking Euclidean curves in the plane. However, one of the main reasons why this complexity is possible may be due to disturbance control indoors. In general, reference trajectories are not more complex than straight lines, curved arcs, or smooth curves. Regarding the number of vehicles, it is evident that the most frequent numbers are 4 or 3, with  \citep{BÃrhaug2011493} being an outlier, where the actual fleet consists of only one boat, and two virtual ones are used to complete the validation set-up.

It is noteworthy to highlight the presence of 5 articles implementing a circulating control. Although this application faces different challenges compared to following a predefined trajectory, it offers a wide range of real-life applications, such as maritime security, military applications, or search and rescue missions. A study addressing validation closely related to some of these applications is \citep{Hu2022}, where three vessels manage to reach, encircle, and rotate around a virtual target. On the other hand, study \citep{Hu2021} proposes a control system capable of reaching one or more targets and maintaining a square formation around them while escorting.

Another essential element in validating control algorithms is the vessels themselves. While these can be classified in various ways, this review will focus on the characteristics outlined in Table \ref{fig: data_model_experimental}. It is observed that the number of boat models is even more limited than the number of articles. Some of the most recurring vessels are the Huster-12 and 0.3 models, each mentioned in three articles. Both vessels feature similar hull designs but differ in size; the former being a medium-sized model, suitable for tackling outdoor environments such as lakes or rivers, while the latter is a small model used for indoor testing with force limitations.

From a design or hull shape standpoint, most vessels have a very similar design, primarily monohull. However, the case of the DUCKLINGS presented in \citep{Ringback2021}, which are catamarans, stands out. Although both designs have their advantages and disadvantages, some of the main differences lie in their stability against disturbances, speed, and cargo capacity. It could be argued that catamarans are more suitable for applications requiring strength and calm waters, while monohulls are more commonly used in applications demanding speed and facing stronger currents.

While not all documents provide a detailed technical description of the vehicles, eight of the eleven models mentioned in the Table  \ref{fig: data_model_experimental} have medium size and weight, allowing them to perform tasks outdoors but with limitations in cargo and force. In contrast, the Odin and Frigg models presented in study \citep{Eek2021171} are large-scale, capable of carrying up to 2 persons and performing maneuvers in open sea thanks to their waterjet propulsion system. A similar case is observed with the Jellyfish used in \citep{Li2020}, which could also transport at least one person and, due to their design, face adverse conditions. It is important to note that these models were the only ones capable of validating their algorithms in open sea and with global route planning.

When delving into the architecture of different vessels, four main components can be identified: power supply, processor, actuators, and sensors. Since experimental validations are often short-term exercises, the characteristics of the power supply are not frequently highlighted; however, in study \citep{Arrichiello201273}, a service duration of approximately 12 hours is specified. This service duration has a direct relationship with the effort and speed of the vessels. However, in most of the reviewed documents, the reference speeds do not exceed 5 m/s, except for those using the Huster-12 model.

In underactuated ASVs the most commonly used actuators are a combination of two thrusters, allowing for differential movement, or a combination of rudder and thruster. In comparison, the latter, although simpler, presents greater constraints regarding rotation and movement. On the other hand, the most implemented sensors are IMUs and GPS, followed by sensors such as cameras, radars, or Light Detection and Ranging (LIDAR), with the latter being widely used for obstacle detection.

To conclude this section, it is worth mentioning that although communication among ASVs has been the subject of extensive research, there is a clear deficiency in experimental validation. While documents with numerical validations propose the use of 7 or more agents, as is the case in \citep{Liang2020Distributed}, experimental validations are limited to fleets of 4 or fewer, using communication techniques such as Wireless or Zigbee. Although these technologies have high consumption, they offer sufficient bandwidth to transmit the desired information among the different vessels, without representing a significant restriction in the validation process.

Similarly, it is evident that the limited number of agents in the experimental validations has not been sufficient to assess the impact of communication topologies on fleet controller performance and to identify potential bottlenecks such as medium access or broadcast interference. An example of this is the \citep{Liu2020Collective} study, where three agents and one leader were implemented with 433M Wireless communication among neighbors or with the leader, within a range of 125 meters. The author acknowledges the importance of evaluating the impact of message loss or connection loss in future work.

Regarding communication topologies, there is a clear preference for neighbor-to-neighbor communication in experimental validation. However, given the number of agents in the fleets, broadcast communication among all agents could be a viable option, improving message uniformity, reducing delivery times, and covering a larger area. An example of this is the work by \citep{Tang2023}, where a fleet of three agents navigates through a narrow channel, obtaining distance measurements via Wireless communication. Although this experiment was conducted indoors, a potential research direction would be to evaluate the same system with more agents to understand the limits of broadcast communication and assess the impact of interference on fleet controllers.

\subsection{RQ.3: Which coordinated control methods are used in each scenario, taking into account the influence of method configuration and communication topology?} \label{subsec:RQ3}
In addressing the coordinated control methods used in ASVs, this review dissects the prevalent coordination strategies in what follows (see Table \ref{fig: formation_control_strategies}):
\begin{table*}[b]
\caption{Classification of formation control strategies}
	\centering
		\begin{tabular}{cc}\toprule
                     & References \\ \midrule
 Leader-follower & \citep{chen2020}, \citep{Fu2018Bioinspired}, \citep{Fu2018Formation}, \citep{Ghommam2018}, \citep{He2021}, \citep{He2022}, \citep{Huang2021Robust}, \citep{Li2019}, \citep{Liang2020Adaptive}, \citep{Lin2022Neural}, \citep{Liu2022RobustFuzzy}, \citep{Sun2018}, \citep{Yue2019}, \citep{Yuhan2022}, \citep{Zhang2021NovelEventTriggered}, \citep{Zhang2022AdaptiveDistributed}, \citep{Zhang2022RobustAdaptive} \\
 &\\
 
 Flocking &  \citep{do2016}, \citep{Jiao2014}, \citep{Liang2021SwarmVelocity}, \citep{Liang2020Distributed}, \citep{liang2019SwarmControl}, \citep{Liu2016Coordinated}, \citep{Liu2017Saturated}, \citep{Panagou2013}, \citep{Peng2012NeuralAdaptive}, \citep{Qin2017}, \citep{Zhuang2019}\\
 &\\
 Formation-control & \citep{BabiÄ‡2016361}, \citep{Bai20205015}, \citep{BÃrhaug2011493}, \citep{chen2019}, \citep{Chen2015}, \citep{chen2022_hierarchical}, \citep{Chen2022}, \citep{Chen2022reinforcement}, \citep{Chen2015Global}, \citep{dai2022}, \citep{Do2012}, \citep{dong2021}, \citep{Dong2010}, \citep{dong2022}, \citep{Eek2021171}, \citep{Fahimi2008}, \citep{Fan2017}, \citep{Fu2022}, \citep{Gao2023Fixed}, \citep{Gao2021} \\
 & \citep{Ghommam2020}, \citep{Ghommam2009}, \citep{Gu2019Antidisturbance}, \citep{Guo2021}, \citep{Han2022}, \citep{hu2023spontaneous}, \citep{Huang2021Adaptive}, \citep{Huang2019Improved}, \citep{Huang2022}, \citep{Jiang2022NonfragileFormationControl}, \citep{Jiang2022Sliding}, \citep{Jiang2022Cooperative}, \citep{Jin2022}, \citep{Jin2016}, \citep{Li2018Adaptive}, \citep{Li2018Finite}, \citep{Li2020}, \citep{Liang2022Secure}, \citep{Liu2021RobustEvent} \\
 & \citep{Liu2022Distributed}, \citep{Liu2022Formation}, \citep{Lu2018}, \citep{Lv2021}, \citep{Lv2022}, \citep{Ma2019}, \citep{Meng2012}, \citep{Mu2020}, \citep{Ning2022}, \citep{Park2022}, \citep{Peng2011RobustAdaptive}, \citep{Peng2013AdaptiveDynamic}, \citep{Peng2012RobustLeader}, \citep{Riahifard2020}, \citep{Shojaei2015}, \citep{Song2022}, \citep{Tang2023}, \\
 & \citep{Wang2019}, \citep{Wang2021}, \citep{Wang2012}, \citep{Wu2021}, \citep{Xia2021}, \citep{Xia2020}, \citep{Xiong2022}, \citep{Ye2021}, \citep{Zhang2020FormationControlMultiple}, \citep{Zhang2022EventTriggered}, \citep{Zhang2022Bearing}, \citep{Zhang2019FormationControlAutonomous}, \citep{Zhao2021}, \citep{Zhou2020} \\
 &\\
 Formation-maneuver &  \citep{Arrichiello201273}, \citep{bishop2012}, \citep{fahimi2007SlidingMode}, \citep{Fahimi2007Nonlinear}, \citep{Gu2019Distributed}, \citep{Gu2022}, \citep{Hao2022}, \citep{kim2021}, \citep{Liang2019Novel}, \citep{Liu2020Collective}, \citep{Liu2022Scanning}, \citep{Lv2023}, \citep{Park2021}, \citep{Peng2019PathGuided}, \citep{Sun2020}, \citep{Sun2022}, \citep{Wang2022}, \citep{Wu2022}, \citep{Wu2023}\\
 & \citep{Xia2022}, \citep{Zhang2021LVS}, \citep{Zhou2022}\\
 &\\
 Containment-control & \citep{Hu2022}, \citep{Hu2021}, \citep{hu2022bearing}, \citep{Ringback2021}, \citep{Yu2019}\\
\bottomrule
	\end{tabular}
\label{fig: formation_control_strategies}
\end{table*}

\begin{table*}[BP]
\caption{Coordinated control method}
	\centering
		\begin{tabular}{cc}\toprule
                     & References \\ \midrule
 TRCC &  \citep{Arrichiello201273}, \citep{Bai20205015}, \citep{bishop2012}, \citep{chen2019}, \citep{chen2022_hierarchical}, \citep{do2016}, \citep{Dong2010}, \citep{dong2022}, \citep{Han2022}, \citep{Jiang2022NonfragileFormationControl}, \citep{Jiang2022Sliding}, \citep{Li2018Adaptive}, \citep{Li2020}, \citep{Liang2019Novel},  \citep{Liang2020Distributed}, \citep{Liang2022Secure}, \citep{liang2019SwarmControl}, \citep{Liu2020Collective}, \citep{Liu2017Saturated},\\
 & \citep{Lv2021}, \citep{Ning2022}, \citep{Peng2011RobustAdaptive}, \citep{Peng2013AdaptiveDynamic}, \citep{Peng2019PathGuided}, \citep{Song2022}, \citep{Wang2022}, \citep{Wang2019}, \citep{Wang2012}, \citep{Yu2019}\\
 &\\
 PACC & \citep{BÃrhaug2011493}, \citep{Chen2015}, \citep{Chen2022reinforcement}, \citep{dai2022}, \citep{Do2012}, \citep{Eek2021171}, \citep{Fan2017}, \citep{Fu2022}, \citep{Fu2018Bioinspired}, \citep{Fu2018Formation}, \citep{Ghommam2020}, \citep{Ghommam2009}, \citep{Gu2019Antidisturbance}, \citep{Gu2019Distributed}, \citep{Gu2022}, \citep{hu2023spontaneous}, \citep{Huang2022}, \citep{Liang2021SwarmVelocity}, \citep{Liu2016Coordinated}, \citep{Lv2023}, \citep{Lv2022} \\
 & \citep{Park2022}, \citep{Park2021}, \citep{Qin2017}, \citep{Sun2018}, \citep{Sun2022}, \citep{Wu2022}, \citep{Wu2021}, \citep{Wu2023}, \citep{Xia2021}, \citep{Xia2022}, \citep{Xia2020}, \citep{Xiong2022}, \citep{Ye2021}, \citep{Yue2019}, \cite{Yuhan2022}, \citep{Zhang2022EventTriggered}, \citep{Zhang2021LVS}, \citep{Zhang2021NovelEventTriggered}, \\
 & \citep{Zhang2022AdaptiveDistributed}, \citep{Zhang2022RobustAdaptive}, \citep{Zhang2019FormationControlAutonomous}, \citep{Zhao2021}, \citep{Zhou2020}, \\
 &\\
 TACC & \citep{Chen2022}, \citep{chen2020}, \citep{dong2021}, \citep{fahimi2007SlidingMode}, \citep{Fahimi2008}, \citep{Fahimi2007Nonlinear}, \citep{Gao2021}, \citep{Gao2023Fixed}, \citep{Ghommam2018}, \citep{Guo2021}, \citep{Hao2022}, \citep{He2021}, \citep{He2022}, \citep{Hu2021}, \cite{Huang2021Adaptive}, \citep{Huang2019Improved}, \citep{Huang2021Robust}, \citep{Jiang2022Cooperative}, \citep{Jiao2014}, \citep{Jin2022}, \citep{Jin2016}, \citep{kim2021}, \\
 &  \citep{Li2019}, \citep{Li2018Finite}, \citep{Liang2020Adaptive}, \citep{Lin2022Neural}, \citep{Liu2021RobustEvent}, \citep{Liu2022RobustFuzzy}, \citep{Liu2022Distributed}, \citep{Liu2022Scanning}, \citep{Liu2022Formation}, \citep{Lu2018}, \citep{Ma2019}, \citep{Meng2012}, \citep{Mu2020}, \citep{Peng2012NeuralAdaptive}, \citep{Peng2012RobustLeader}, \citep{Riahifard2020}, \citep{Shojaei2015}, \citep{Sun2020}, \citep{Tang2023}, \\
 & \citep{Wang2021}, \citep{Zhang2022Bearing}, \citep{Zhou2022}, \citep{Zhuang2019}\\
 &\\
 FSCC &  \citep{BabiÄ‡2016361}, \citep{Chen2015Global}, \citep{Hu2022}, \citep{hu2022bearing}, \citep{Panagou2013}, \citep{Ringback2021}, \citep{Zhang2020FormationControlMultiple}\\
\bottomrule
	\end{tabular}
\label{fig: control_method}
\end{table*}

\begin{itemize}
    \item Leader-Follower: This approach involves a designated leader ASV guiding the movement of follower ASVs. The followers maintain a predefined position and orientation relative to the leader, adapting to its changes in trajectory. This method, evidenced by multiple references is widely used due to its simplicity and effectiveness in structured environments.
    \item Flocking: Inspired by the natural behavior of birds, flocking algorithms enable ASVs to move as a cohesive group without a fixed formation, adjusting their paths based on the position and velocity of their neighbors. This method, supported by many studies, offers flexibility and scalability, making it suitable for scouting and surveillance tasks. The fundamental rules of flocking, which promote this coordinated behavior, are threefold:

- separation: avoid crowding by maintaining a comfortable distance from neighbors;\\
- alignment: move in roughly the same direction and speed as nearby neighbors;\\
- cohesion: stay close to neighbors to move as a group.
    \item Formation Control: This technique allows multiple ASVs to maintain a predefined geometric shape while moving. It is characterized by its robustness and adaptability to dynamic environments, as seen in extensive research. The large number of references highlights its popularity and broad applicability in tasks requiring coordinated precision.
    \item Formation Maneuver: Extending beyond static formations, this method focuses on the coordinated transition between different formations while considering the kinematic and dynamic constraints of ASVs. References illustrate its application in complex navigational tasks, where adaptability and precision are crucial.
    \item Containment Control: In this strategy, a subset of ASVs, known as leaders, guide the overall group within certain boundaries, while the rest, the followers, ensure they remain within the leaders' formation. This method, although less cited, is useful for tasks requiring the ASVs to operate within constrained areas.
\end{itemize}

Another important aspect considered in the review is the classification of existing coordinated control methods for ASVs based on various types of reference signals. Table \ref{fig: control_method} provides a comprehensive overview, delineating these methods into four primary categories:
\begin{table*}[H]
\caption{Communication topologies}
	\centering
		\begin{tabular}{cc}\toprule
                     & References   \\ \midrule
 All to all &  \citep{Liu2020Collective}, \citep{Shojaei2015}, \citep{Tang2023}, \citep{Zhao2021}\\
  &\\
 Leader &  \citep{chen2020}, \citep{Chen2022reinforcement}, \citep{dong2021}, \citep{dong2022}, \citep{Fu2018Bioinspired}, \citep{Fu2018Formation}, \citep{Ghommam2018}, \citep{Ghommam2020}, \citep{Guo2021}, \citep{hu2022bearing}, \citep{Huang2022}, \citep{Huang2021Robust}, \citep{Jiao2014}, \citep{Jin2022}, \citep{Jin2016}, \citep{kim2021}, \citep{Li2019}, \citep{Li2018Adaptive}, \citep{Lin2022Neural}, \citep{Liu2022RobustFuzzy}, \citep{Lu2018},\\
 & \citep{Panagou2013}, \citep{Peng2011RobustAdaptive}, \citep{Peng2013AdaptiveDynamic}, \citep{Peng2012RobustLeader}, \citep{Riahifard2020}, \citep{Sun2020}, \citep{Sun2018}, \citep{Wang2021}, \citep{Wang2012}, \citep{Wu2022}, \citep{Zhang2019FormationControlAutonomous},\citep{Zhang2021LVS}, \citep{Zhang2021NovelEventTriggered}, \citep{Zhou2020} \\
  &\\
 Neighbour &  \citep{Bai20205015}, \citep{BÃrhaug2011493}, \citep{chen2019}, \citep{Chen2015}, \citep{chen2022_hierarchical}, \citep{Chen2015Global}, \citep{do2016}, \citep{Do2012}, \citep{Dong2010}, \citep{fahimi2007SlidingMode}, \citep{Fahimi2008}, \citep{Fahimi2007Nonlinear}, \citep{Fan2017}, \citep{Ghommam2009}, \citep{Gu2019Antidisturbance}, \citep{Gu2019Distributed}, \citep{Gu2022}, \citep{Han2022}, \citep{Hao2022}, \citep{He2022}, \\
  & \citep{Hu2022}, \citep{Hu2021}, \citep{hu2023spontaneous}, \citep{Huang2022}, \citep{Jiang2022NonfragileFormationControl}, \citep{Jiang2022Sliding}, \citep{Li2019}, \citep{Li2018Adaptive}, \citep{Li2018Finite}, \citep{Liang2019Novel}, \citep{Liang2021SwarmVelocity}, \citep{Liang2020Distributed}, \citep{Liang2022Secure}, \citep{Liu2016Coordinated}, \citep{Liu2017Saturated}, \citep{Liu2022Distributed}, \citep{Liu2022Scanning}, \citep{Liu2022Formation}, \citep{Lv2021}\\ 
 & \citep{Lv2023}, \citep{Ma2019}, \citep{Meng2012}, \citep{Mu2020}, \citep{Ning2022}, \citep{Park2022}, \citep{Park2021}, \citep{Peng2012NeuralAdaptive}, \citep{Peng2019PathGuided}, \citep{Qin2017}, \citep{Ringback2021}, \citep{Wang2022}, \citep{Wang2019}, \citep{Wu2022}, \citep{Wu2021}, \citep{Wu2023}, \citep{Xia2021}, \citep{Xia2022}, \citep{Xia2020},\\
 &   \citep{Xiong2022}, \citep{Ye2021}, \citep{Yu2019}, \citep{Yue2019}, \citep{Yuhan2022}, \citep{Zhang2022EventTriggered}, \citep{Zhang2022Bearing}, \citep{Zhang2022RobustAdaptive}\\
  &\\
 Not-specify & \citep{Arrichiello201273}, \citep{BabiÄ‡2016361}, \citep{bishop2012}, \citep{Eek2021171}, \citep{Fu2022}, \citep{He2021}, \citep{Huang2019Improved}, \citep{Jiang2022Cooperative}, \citep{Li2020}, \citep{Liang2020Adaptive}, \citep{liang2019SwarmControl}, \citep{Liu2021RobustEvent}, \citep{Sun2022}, \citep{Zhang2020FormationControlMultiple}, \citep{Zhuang2019} \\
  &\\
 Asynchronous & \citep{Chen2022}, \citep{Gao2023Fixed}, \citep{Gao2021}, \citep{Lv2022}, \citep{Song2022}, \citep{Zhang2022AdaptiveDistributed}, \citep{Zhou2022}\\
\bottomrule
	\end{tabular}
\label{fig: communication_topologies}
\end{table*}

\begin{itemize}
    \item Trajectory-Guided Coordinated Control (TRCC): This method focuses on guiding a group of surface vehicles to follow one or multiple time-specific predefined trajectories. The implementation can be centralized, where all vehicles have access to the trajectory data, or distributed, where vehicles rely on exchanging information to maintain formation. In distributed settings, each vehicle gathers data from its neighbors through communication or sensors, necessitating consideration of individual vehicle dynamics and the constraints of communication networks. The significant number of references underlines the popularity and effectiveness of TRCC in ensuring coordinated movement with high temporal and spatial fidelity.
    \item Path-Guided Coordinated Control (PACC): The aim here is to navigate a group of surface vehicles along predefined paths in a specific formation and at a set speed. This approach can be split into two main types: one where the path information is fully known and another where it's partially known. In the fully known scenario, control laws ensure each vehicle aligns with its path, and a distributed law updates path variables to coordinate the formation, including tasks like synchronization and containment. This often requires centralized planning for each vehicle's path. In the partially known scenario, the lack of complete path information for every vehicle necessitates a distributed control approach. The diversity of references in this category reflects the adaptability of PACC to various operational contexts, from structured environments to more dynamic, unpredictable settings.
    \item Target-Guided Coordinated Control (TACC): This strategy involves aligning one or more surface vehicles into a formation to track one or multiple targets, differing from trajectory tracking by not having predefined velocity and acceleration data. Instead, target details such as speed and direction might need to be estimated collaboratively due to sensing or communication limits, ensuring consensus among the vehicles. The references under TACC demonstrate the method's reliance on robust estimation and control algorithms to compensate for the uncertainty and variability inherent in tracking moving targets.
    \item Formation Stabilization Cooperative Control (FSCC): Unlike other methods that may involve following a leader or a specific path, FSCC's sole purpose is to position ASVs into a predetermined formation and maintain their positions stably without further movement. It is currently the least encountered method in the literature
\end{itemize}

To effectively navigate the complexities of coordinated control in ASVs, it is crucial to understand the underlying communication aspects. On the one hand, communication topologies determine how individual ASVs within a fleet share information and make collective decisions, significantly affecting the system's robustness, flexibility, and efficiency. Table \ref{fig: communication_topologies} classifies the primary communication topologies identified in this research, from fully connected networks where all units communicate with each other, to more structured systems such as leader-based and neighbor-based systems.

On the other hand, a clear distinction emerges between synchronous and asynchronous communication. Synchronous methods, known for their coordinated timing in communication and operations, are predominant. Within this category, the 'neighbour' topology is particularly notable, highlighting a decentralized approach where each ASV exchanges information locally with adjacent vehicles. This strategy enhances scalability and adaptability, which are vital in dynamic maritime settings. Closely following is the 'leader' topology, employing a more hierarchical structure to provide explicit guidance and coordination. Furthermore, the 'asynchronous' category includes the aperiodic communication methods, which, are less comparatively studied in the literature but can result in efficiency in the bandwidth usage and prevent delays by reducing the communication rates. The limited references to asynchronous topologies indicate an emerging interest, suggesting potential avenues for future research and development in ASV coordinated control systems.

\begin{table*}[H]
\caption{Additional characteristics extracted from each article}
	\centering
		\scriptsize{
		\begin{tabular}{{>{\centering\arraybackslash}m{2.165em}
                >{\centering\arraybackslash}m{16.15em}
                >{\centering\arraybackslash}m{6.50em}
                >{\centering\arraybackslash}m{6.15em}
                >{\centering\arraybackslash}m{6.15em}
                >{\centering\arraybackslash}m{6.15em}
                >{\centering\arraybackslash}m{6.15em}}}\toprule
                  Ref  & Controller & Information exchanged & Own information & Coordinated control method & Architecture & Transmission type\\ \midrule 
                  \multicolumn{7}{c}{\textbf{Leader-follower}} \\ \hline

                \citep{chen2020} & Sliding-modes & p-a & p-a-s & TACC & Distributed & Synchronous \\
                \citep{Fu2018Bioinspired} & Sliding-modes & p-s & p-s & PACC & Distributed & Synchronous \\
                 \citep{Fu2018Formation} & Backstepping & p-s & p-s & PACC & Distributed & Synchronous \\
                 \citep{Ghommam2018} & BLF and Adaptative   Backstepping & p-a & p-a-s & TACC & Distributed & Synchronous \\
                 \citep{He2021} & BLF and Backstepping & p & p-a-s & TACC & Decentralized & Synchronous \\
                 \citep{He2022} & Neural-networks & p & p-a-s & TACC & Decentralized & Synchronous \\
                 \citep{Huang2021Robust} & Robust & p-a & p-a & TACC & Distributed & Synchronous \\
                 \citep{Li2019} & Feedback linearization & p-s & p-a-s & TACC & Distributed & Synchronous \\
                 \citep{Liang2020Adaptive} & Adaptive & a-s & p-a-s & TACC & Distributed & Synchronous \\
                 \citep{Lin2022Neural} & BLF & p & p-a & TACC & Distributed & Synchronous \\
                 \citep{Liu2022RobustFuzzy} & Fuzzy & p-a & p-a & TACC & Distributed & Synchronous \\
                 \citep{Sun2018} & Sliding-modes & p-a & p-a-s & PACC & Decentralized & Synchronous \\
                 \citep{Yue2019} & Robust & p-a-s & p-a-s & PACC & Distributed & Synchronous \\
                 \citep{Yuhan2022} & Adaptive & p-a-s & p-a-s & PACC & Distributed & synchronous \\
                 \citep{Zhang2021NovelEventTriggered} & Adaptive & p-a-s & p-a-s & PACC & Distributed & Synchronous \\
                 \citep{Zhang2022AdaptiveDistributed} & Adaptive & p-a-s & p-a-s & PACC & Distributed & Asynchronous \\
                 \citep{Zhang2022RobustAdaptive} & Adaptive & p-a-s & p-a-s & PACC & Distributed & Synchronous \\

             \midrule 
                  \multicolumn{7}{c}{\textbf{Flocking}} \\ \hline
            
              \citep{do2016} & Backstepping & p-a-s & p-a-s & TRCC & Distributed & Synchronous \\
             \citep{Jiao2014} & Sliding-modes & p & p-a & TACC & Distributed & Synchronous \\
             \citep{Liang2021SwarmVelocity} & BLF & p & p-a-s & PACC & Distributed & Synchronous \\
             \citep{Liang2020Distributed} & Neural-networks & p & p-a-s & TRCC & Distributed & Synchronous \\
             \citep{liang2019SwarmControl} & Robust & p & p-a-s & TRCC & Distributed & Synchronous \\
             \citep{Liu2016Coordinated} & Backstepping & p & p-a-s & PACC & Distributed & Synchronous \\
             \citep{Liu2017Saturated} & Neural-networks & a-s & p-a & TRCC & Distributed & Synchronous \\
             \citep{Panagou2013} & Backstepping & p-a & p-a-s & FSCC & Decentralized & Synchronous \\
             \citep{Peng2012NeuralAdaptive} & Adaptive & p-a-s & p-a-s & TACC & Distributed & Synchronous \\
             \citep{Qin2017} & Sliding-modes & p & p-a & PACC & Decentralized & Synchronous \\
             \citep{Zhuang2019} & Sliding-modes & N/S & p-a & TACC & Decentralized & Synchronous \\

             \midrule
                \multicolumn{7}{c}{\textbf{Formation-maneuver}} \\ \hline
            
              \citep{Arrichiello201273} & Null-Space-based Behavioral & p & p & TRCC & Distributed & Synchronous \\
             \citep{bishop2012} & Redundant-Manipulator & p-a-s & p-a-s & TRCC & Decentralized & Synchronous \\
             \citep{fahimi2007SlidingMode} & Sliding-modes & p-a & p-a & TACC & Decentralized & Synchronous \\
             \citep{Fahimi2007Nonlinear} & MPC & p-a & p-a & TACC & Decentralized & Synchronous \\
             \citep{Gu2019Distributed} & Backstepping & p & p-a-s & PACC & Distributed & Synchronous \\
             \citep{Gu2022} & Adaptive & p & p-a-s & PACC & Distributed & Synchronous \\
             \citep{Hao2022} & Backstepping & p-a & p-a-s & TACC & Distributed & Synchronous \\
             \citep{Hu2022} & PID & p & p-a-s & FSCC & Distributed & Synchronous \\
             \citep{Hu2021} & Adaptive & p-a-s & p-a-s & TACC & Distributed & Synchronous \\
             \citep{hu2022bearing} & Backstepping & a & a & FSCC & Distributed & Synchronous \\
             \citep{kim2021} & PID & p & p-a & TACC & Distributed & Synchronous \\
             \citep{Liang2019Novel} & Robust & p & p-a & TRCC & Distributed & Synchronous \\
             \citep{Liu2020Collective} & PID & p-s & p-a-s & TRCC & Distributed & Synchronous \\
             \citep{Liu2022Scanning} & MPC & p-a & p-a-s & TACC & Distributed & Synchronous \\
             \citep{Lv2023} & MPC & p-a & p-a-s & PACC & Distributed & Synchronous \\
             \citep{Park2021} & Backstepping & p-a-s & p-a-s & PACC & Distributed & Synchronous \\
             \citep{Peng2019PathGuided} & Adaptive & p-a & p-a-s & TRCC & Distributed & Synchronous \\
             \citep{Ringback2021} & PID & p & p-a-s & FSCC & Distributed & Synchronous \\
             \citep{Sun2020} & MPC & gt & p-a-s & TACC & Distributed & Synchronous \\
             \citep{Sun2022} & Sliding-modes & p-a & p-a-s & PACC & Distributed & Synchronous \\
             \citep{Wang2022} & Robust & p-a & p-a-s & TRCC & Distributed & Synchronous \\
             \citep{Wu2022} & BLF & p-a-s & p-a-s & PACC & Distributed & Synchronous \\
             \citep{Wu2023} & BLF & p-a & p-a & PACC & Distributed & Synchronous \\
             \citep{Xia2022} & Backstepping & p-a & p-a & PACC & Distributed & Synchronous \\
             \citep{Yu2019} & Extended-state-observer & p-a-s & p-a-s & TRCC & Distributed & Synchronous \\
             \citep{Zhang2021LVS} & Adaptive & p-a & p-a & PACC & Distributed & Synchronous \\
             \citep{Zhou2022} & Adaptive & p-a-s & p-a-s & TACC & Distributed & Asynchronous \\
        \bottomrule
	\end{tabular}}
\label{fig: communication1}
\end{table*}

\begin{table*}[H]
	\centering
		\scriptsize{
		\begin{tabular}{{>{\centering\arraybackslash}m{2.165em}
                >{\centering\arraybackslash}m{16.15em}
                >{\centering\arraybackslash}m{6.50em}
                >{\centering\arraybackslash}m{6.15em}
                >{\centering\arraybackslash}m{6.15em}
                >{\centering\arraybackslash}m{6.15em}
                >{\centering\arraybackslash}m{6.15em}}}\toprule
                  Ref  & Controller & Information transmitted & Own information & Coordinated control method & Architecture & Transmission type\\ \midrule 
                   \multicolumn{7}{c}{\textbf{Formation-control}} \\ \hline
                   
                 \citep{BabiÄ‡2016361} & Adaptive & p & p & FSCC & Decentralized & Synchronous \\
                 \citep{Bai20205015} & Sliding-modes & p & p-a-s & TRCC & Distributed & Synchronous \\
                 \citep{BÃrhaug2011493} & Feedback linearization & p & p-a-s & PACC & Decentralized & Synchronous \\
                 \citep{chen2019} & Adaptive & p & p & TRCC & Centralized & Synchronous \\
                 \citep{Chen2015} & Backstepping & p-a-s & p-a-s & PACC & Decentralized & Synchronous \\
                 \citep{chen2022_hierarchical} & Backstepping & p-a-s & p-a-s & TRCC & Centralized & Synchronous \\
                 \citep{Chen2022} & Adaptive & p-a-s & p-a-s & TACC & Distributed & Asynchronous \\
                 \citep{Chen2022reinforcement} & Sliding-modes & p-a & p-a & PACC & Decentralized & Synchronous \\
                 \citep{Chen2015Global} & Backstepping & p-a-s & p-a-s & FSCC & Decentralized & Synchronous \\
                 \citep{dai2022} & BLF and Backstepping & p-a & p-a & PACC & Decentralized & Synchronous \\
                 \citep{Do2012} & Backstepping & p-a & p-a-s & PACC & Distributed & Synchronous \\
                 \citep{dong2021} & Backstepping & p-a-s & p-a-s & TACC & Distributed & Synchronous \\
                 \citep{Dong2010} & Lyapunov-based & p-a-s & p-a-s & TRCC & Decentralized & Synchronous \\
                 \citep{dong2022} & Sliding-modes & p-a-s & p-a-s & TRCC & Distributed & Synchronous \\
                 \citep{Eek2021171} & Sliding-modes & p-a & p-a & PACC & Centralized & Synchronous \\
                 \citep{Fahimi2008} & Backstepping & p-a & p-a & TACC & Decentralized & Synchronous \\
                 \citep{Fan2017} & MPC & p-a & p-a & PACC & Distributed & Synchronous \\
                 \citep{Fu2022} & Finite-time-homogenity & p-a & p-a & PACC & Decentralized & Synchronous \\
                 \citep{Gao2023Fixed} & Fixed-time & p-a & p-a & TACC & Distributed & Asynchronous \\
                 \citep{Gao2021} & Backstepping & p-a & p-a & TACC & Distributed & Asynchronous \\
                 \citep{Ghommam2020} & Neural-networks & p-a & p-a & PACC & Distributed & Synchronous \\
                 \citep{Ghommam2009} & Backstepping & a & p-a & PACC & Distributed & Synchronous \\
                 \citep{Gu2019Antidisturbance} & Backstepping & p & p-a-s & PACC & Distributed & Synchronous \\
                 \citep{Guo2021} & BLF & p-a-s & p-a-s & TACC & Distributed & Synchronous \\
                 \citep{Han2022} & Adaptive & p & p-a-s & TRCC & Distributed & Synchronous \\
                 \citep{hu2023spontaneous} & Backstepping & gt & p-a & PACC & Distributed & Synchronous \\
                 \citep{Huang2021Adaptive} & Adaptive & p-a & p-a-s & TACC & Decentralized & Synchronous \\
                 \citep{Huang2019Improved} & Adaptive & N/S & p-a-s & TACC & Decentralized & Synchronous \\
                 \citep{Huang2022} & Backstepping & p-a & p-a & PACC & Distributed & Synchronous \\
                 \citep{Jiang2022NonfragileFormationControl} & Lyapunov-based & p-s & a-s & TRCC & Distributed & Synchronous \\
                 \citep{Jiang2022Sliding} & Sliding-modes & p-s & p-s & TRCC & Distributed & Synchronous \\
                 \citep{Jiang2022Cooperative} & fuzzy & p-a & p-a & TACC & Distributed & Synchronous \\
                 \citep{Jin2022} & Reinforcement   Learning & p-a & p-a & TACC & Distributed & Synchronous \\
                 \citep{Jin2016} & BLF & p-a & p-a & TACC & Distributed & Synchronous \\
                 \citep{Li2018Adaptive} & Adaptive & p & p-a & TRCC & Distributed & Synchronous \\
                 \citep{Li2018Finite} & Sliding-modes & p & p-a & TACC & Distributed & Synchronous \\
                 \citep{Li2020} & Feedback linearization & p & p-a-s & TRCC & Distributed & Synchronous \\
                 \citep{Liang2022Secure} & Feedback linearization & p & p-a-s & TRCC & Distributed & Synchronous \\
                 \citep{Liu2021RobustEvent} & Robust & p-a-s & p-a-s & TACC & Distributed & Synchronous \\
                 \citep{Liu2022Distributed} & MPC & p-a & p-a & TACC & Distributed & Synchronous \\
                 \citep{Liu2022Formation} & Sliding-modes & p-a-s & p-a-s & TACC & Distributed & Synchronous \\
                 \citep{Lu2018} & Neural-networks & p-a & p-a-s & TACC & Decentralized & Synchronous \\
                 \citep{Lv2021} & MPC & p-a & p-a-s & TRCC & Distributed & Synchronous \\
                 \citep{Lv2022} & Backstepping & gt & p-a-s & PACC & Distributed & Asynchronous \\
                 \citep{Ma2019} & Adaptive & p-a & p-a-s & TACC & Distributed & Synchronous \\
                 \citep{Meng2012} & Sliding-modes & p-a & p-a-s & TACC & Distributed & Synchronous \\
                 \citep{Mu2020} & Adaptive & p-a & p-a & TACC & Distributed & Synchronous \\
                 \citep{Ning2022} & Adaptive & p-a & p-a-s & TRCC & Distributed & Synchronous \\
                 \citep{Park2022} & Neural-networks & p-a & p-a-s & PACC & Distributed & Synchronous \\
                 \citep{Peng2011RobustAdaptive} & Adaptive & p-a & p-a-s & TRCC & Decentralized & Synchronous \\
                 \citep{Peng2013AdaptiveDynamic} & Adaptive & p-a & p-a-s & TRCC & Decentralized & Synchronous \\
                 \citep{Peng2012RobustLeader} & Backstepping & p-a & p-a-s & TACC & Decentralized & Synchronous \\
                 \citep{Riahifard2020} & Adaptive & p-a & p-a & TACC & Distributed & Synchronous \\
                 \citep{Shojaei2015} & Adaptive & p-a-s & p-a-s & TACC & Decentralized & Synchronous \\
                 \citep{Song2022} & Sliding-modes & ca & p-a-s & TRCC & Distributed & Asynchronous \\
                 \citep{Tang2023} & PID & p-a & p-a & TACC & Distributed & Synchronous \\
                 \citep{Wang2019} & BLF & p & p-a-s & TRCC & Distributed & Synchronous \\
                 \citep{Wang2021} & BLF & p-a & p-a-s & TACC & Distributed & Synchronous \\
                 \citep{Wang2012} & Adaptive & p-a & p-a-s & TRCC & Distributed & Synchronous \\
                 \citep{Wu2021} & Adaptive & p-a-s & p-a-s & PACC & Distributed & Synchronous \\
                 \citep{Xia2021} & Backstepping & p-a & p-a & PACC & Distributed & Synchronous \\
                 \citep{Xia2020} & BLF & p-a & p-a & PACC & Distributed & Synchronous \\
                 \citep{Xiong2022} & BLF & p-a-s & p-a-s & PACC & Distributed & Synchronous \\
                 \citep{Ye2021} & Finite-time-homogeneity & p-a-s & p-a-s & PACC & Distributed & Synchronous \\
                 \citep{Zhang2022EventTriggered} & Adaptive & p-a-s & p-a-s & PACC & Distributed & Synchronous \\
                 \citep{Zhang2020FormationControlMultiple} & Backstepping & p-a-s & p-a-s & FSCC & Decentralized & Synchronous \\
                 \citep{Zhang2022Bearing} & Backstepping & p-a-s & p-a-s & TACC & Distributed & Synchronous \\
                 \citep{Zhang2019FormationControlAutonomous} & Fuzzy & p & p-a & PACC & Distributed & Synchronous \\
                 \citep{Zhao2021} & Neural-networks & p-a-s & p-a-s & PACC & Centralized & Synchronous \\
                 \citep{Zhou2020} & Adaptive & p-a-s & p-a-s & PACC & Distributed & Synchronous\\
            \bottomrule
	\end{tabular}}
\label{fig: communication2}
\end{table*}

Following the discussion on communication aspects, it is crucial to understand not only the type of information exchanged between them but also the self-knowledge each ASV possesses. In Tables \ref{fig: communication1} and \ref{fig: communication2}, when detailing information, "p" refers to position, "a" to angle, "s" to speed, "ca" to control action, and "gt" stands for guidance-targets, which involves the transmission of objective points. Finally, with the abbreviation "N/S," it is understood to mean "Not Specified."
Only in \citep{hu2023spontaneous} and \citep{Sun2020} do the ASVs exchange guidance-targets.

\begin{figure}[ht!]
    \includegraphics[trim={0cm} {0cm} {0cm} {0cm},clip,width=0.5\textwidth]{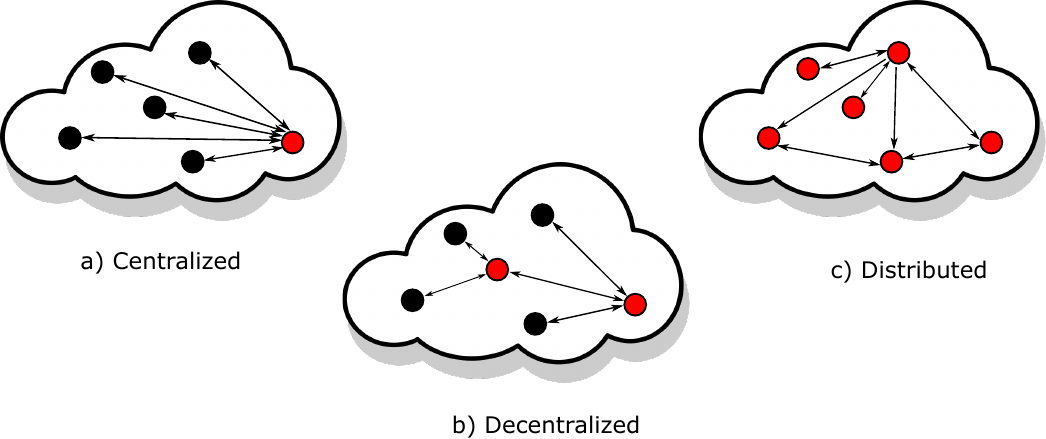}
\caption{Differences between centralized, decentralized and distributed system architecture} \label{fig: diagram_centralized_Distributed}
\end{figure}

In addition, Tables \ref{fig: communication1} and \ref{fig: communication2} compile, for each reference, additional details such as the type of transmission (synchronous or asynchronous), the coordinated control method, and the architecture of the coordinated control system. Figure \ref{fig: diagram_centralized_Distributed} illustrates the distinct characteristics of these three architecture types. In the centralized architecture depicted in Figure \ref{fig: diagram_centralized_Distributed} a, a smart network incorporates a central data fusion hub, allowing each ASV to exchange data with this central hub. Figure \ref{fig: diagram_centralized_Distributed} b outlines the decentralized topology, characterized by the presence of multiple fusion centers that can communicate with each other, while direct communication between ASV is absent. Finally, Figure \ref{fig: diagram_centralized_Distributed} c depicts the distributed system, where ASV must interact directly with each other, sharing their data across a network to collectively gather insights about the entire system.

Another column is dedicated to the associated coordinated control method, as well as the control technique utilized. Furthermore, the tables are divided according to formation control strategies.

\subsection{RQ.4: What machine learning techniques, including deep learning and other neural networks, are used and for what purposes?}\label{subsec:Learning}
To initiate the discussion on the utilization of machine learning, an initial screening was conducted, where only papers employing such techniques were identified. Out of the 125 articles analyzed, 48 employ some form of machine learning technique. Regarding a temporal analysis, it is observed that 33 out of the 48 papers were published between 2020 and 2023. This finding constitutes clear evidence of the current trend towards the implementation of these techniques in fleet management. However, for a more detailed analysis, a discussion will be presented below on the techniques used in each paper and the problems they aim to address within coordinated control.

Table \ref{fig: Machine Learning} presents the 48 identified papers divided according to the coordinated control technique. Homogeneity in their application is evident, although at first glance, it appears that they are more commonly used in formation control. This is mainly due to the number of papers addressing this type of coordinated control. However, it is clear that these techniques contribute to the challenges faced by each type of control, making them scalable for various applications and turning them into one of the most widely used tools today.

Performing a similar analysis to question 1, we have identified the challenges addressed by different machine learning techniques, classifying them into 8 groups:

\begin{itemize}
 \item Model uncertainties: Techniques used to identify or compensate for errors in modeling.

 \item Disturbances: Techniques enabling the calculation or understanding of external disturbances unrelated to the vessel.

 \item Formation: Techniques directly contributing to the performance of coordinated control.

 \item Input Gains: Techniques for calculating actuator gains to achieve desired forces and torques.

 \item Actuator faults: Techniques for detecting and responding to faults in actuators.

 \item Dynamic Control: Techniques contributing to the control of linear and angular velocity per agent.

 \item Estimate: Machine learning techniques for estimating the state of neighboring vessels.

 \item Trajectory: Techniques for calculating or improving reference trajectories.
\end{itemize}

Table \ref{fig: Machine Learning} depicts the problems faced by the machine learning algorithms in each of the articles, along with a brief description of the techniques used and the primary control employed in each case. It is important to note that this table focuses solely on the challenges of machine learning. For a more comprehensive perspective of each article, referring to Table \ref{fig: limitation} is recommended. Furthermore, upon analyzing the most commonly combination with control techniques, the frequency of adaptive control is evident. In most cases, machine learning is employed to update parameters and react to changes in modeling.

\begin{table*}[H]
\caption{Machine learning techniques and their applications}
	\centering
		\scriptsize{
		\begin{tabular}{{>{\centering\arraybackslash}m{2.15em}
                >{\centering\arraybackslash}m{3.15em}
                >{\centering\arraybackslash}m{12.50em}
                >{\centering\arraybackslash}m{16.15em}
                cccccccc}}\toprule
                   \textbf{Ref} & \textbf{Year} & \textbf{Main controller}   & \textbf{ML Tecnique} & 
                   \rotatebox[origin=c]{90}{\parbox{2cm}{\centering\textbf{Model\\ uncertainties}}} &
                   \rotatebox[origin=c]{90}{\textbf{Disturbances}} & 
                   \rotatebox[origin=c]{90}{\textbf{Formation}} &
                   \rotatebox[origin=c]{90}{\textbf{Inputs gains}} &
                   \rotatebox[origin=c]{90}{\parbox{2cm}{\centering\textbf{Actuator\\ faults}}} & 
                   \rotatebox[origin=c]{90}{\parbox{2cm}{\centering\textbf{Dynamic\\ Control}}} & 
                   \rotatebox[origin=c]{90}{\textbf{Estimate}} &
                   \rotatebox[origin=c]{90}{\textbf{Trayectory}}
                   \\ \midrule  
                    \multicolumn{12}{c}{\textbf{leader-follower}} \\ \hline
                    \citep{chen2020} & 2020 & Sliding-modes & RBF-NN & X & X &  &  &  &  &  &  \\
                    \citep{He2021} & 2021 & BLF and Backstepping & NN & X &  &  &  &  &  &  &  \\
                    \citep{He2022} & 2022 & Neural-networks & RBF-NN & X &  & X &  &  &  &  &  \\
                    \citep{Lin2022Neural} & 2022 & BLF & structuring neural   network (SNN) & X & X &  &  &  &  &  &  \\
                   \citep{Liu2022RobustFuzzy} & 2022 & fuzzy & NN - MLP & X & X &  &  &  &  &  &  \\
                    \citep{Zhang2021NovelEventTriggered} & 2021 & Adaptive & RBF-NN & X &  &  &  &  &  &  &  \\
                    \citep{Zhang2022AdaptiveDistributed} & 2022 & Adaptive & NN & X & X &  &  &  &  &  &  \\
                    \citep{Zhang2022EventTriggered} & 2022 & Adaptive & RBF-NN & X & X &  &  &  &  &  &  \\

                     \midrule 
                    \multicolumn{12}{c}{\textbf{flocking}} \\ \hline

                    \citep{Liang2020Distributed} & 2020 & Neural-networks & wavelet neural   network (WNN) & X &  & X &  &  &  &  &  \\
                    \citep{liang2019SwarmControl} & 2019 & Robust & NN & X & X &  &  &  &  &  &  \\
                    \citep{Liu2016Coordinated} & 2016 & Backstepping & predictor-based   neural network & X & X &  &  &  &  &  &  \\
                    \citep{Liu2017Saturated} & 2017 & neural-networks & NN & X & X & X &  &  &  &  &  \\
                    \citep{Peng2012NeuralAdaptive} & 2012 & Adaptive & RBF-NN & X & X &  &  &  &  &  &  \\
                    \midrule 
                    \multicolumn{12}{c}{\textbf{formation-control}} \\ \hline

                    \citep{chen2019} & 2019 & Adaptive & Broad Learning System   (BLS) & X & X &  &  &  &  &  & X \\
                    \citep{Chen2022} & 2022 & Adaptive & NN & X & X &  &  &  &  &  &  \\
                    \citep{Chen2022reinforcement} & 2022 & Sliding-modes & NN and RL &  &  & X &  & X &  &  &  \\
                    \citep{Ghommam2020} & 2020 & Neural-networks & NN & X &  &  &  &  &  &  &  \\
                    \citep{Han2022} & 2022 & Adaptive & RF-NN and MPL & X & X &  &  &  &  &  &  \\
                    \citep{Huang2021Adaptive} & 2021 & Adaptive & NN & X &  &  &  &  &  &  &  \\
                    \citep{Huang2019Improved} & 2019 & Adaptive & NN and MLP & X & X &  &  &  &  &  &  \\
                    \citep{Jin2022} & 2022 & reinforcement   learning & RF-NN & X &  & X &  &  &  &  &  \\
                    \citep{Liu2021RobustEvent} & 2021 & Robust & neural damping method & X &  &  &  &  &  &  &  \\
                    \citep{Lu2018} & 2018 & Neural-networks & RBF-NN and MLP & X &  &  &  &  & X &  &  \\
                    \citep{Ma2019} & 2019 & Adaptive & adaptive neural   networks &  & X &  &  &  &  &  &  \\
                    \citep{Mu2020} & 2020 & Adaptive & MLP & X & X &  &  &  &  &  &  \\
                    \citep{Ning2022} & 2022 & Adaptive & RBF-NN & X & X &  &  &  &  &  &  \\
                    \citep{Park2022} & 2022 & Neural-networks & NN & X & X & X &  &  &  &  &  \\
                    \citep{Peng2011RobustAdaptive} & 2011 & Adaptive & NN & X &  &  &  &  &  & X &  \\
                    \citep{Peng2013AdaptiveDynamic} & 2013 & Adaptive & Nonlinearly   parameterized NNs & X & X &  &  &  &  &  &  \\
                    \citep{Peng2012RobustLeader} & 2012 & Backstepping & Nonlinearly   parameterized NNs & X & X &  &  &  &  &  &  \\
                    \citep{Shojaei2015} & 2015 & Adaptive & RBF-NN & X & X &  &  &  &  &  &  \\
                    \citep{Song2022} & 2022 & Sliding-modes & RBF-NN & X & X &  &  &  &  &  &  \\
                    \citep{Wang2019} & 2019 & BLF & NN & X & X &  & X &  &  &  &  \\
                    \citep{Wang2012} & 2012 & Adaptive & single hidden layer   (SHL) NN & X & X &  &  &  &  &  &  \\
                    \citep{Wu2021} & 2021 & Adaptive & RBF-NN & X & X &  &  &  &  &  &  \\
                    \citep{Xia2021} & 2021 & Backstepping & NN-observer & X & X &  &  &  &  & X &  \\
                    \citep{Xiong2022} & 2022 & BLF & RBF-NN &  & X &  &  &  &  &  &  \\
                    \citep{Zhang2022RobustAdaptive} & 2022 & Adaptive & NN & X &  &  &  &  &  &  &  \\
                    \citep{Zhang2022Bearing} & 2022 & Backstepping & RBF-NN and MLP & X & X &  &  &  &  &  &  \\
                    \citep{Zhao2021} & 2021 & Neural-networks & DRL &  &  & X &  &  &  &  &  \\

                     \midrule 
                    \multicolumn{12}{c}{\textbf{formation-maneuver}} \\ \hline

                    \citep{Gu2022} & 2022 & Adaptive & RBF-NN & X &  &  & X &  &  &  &  \\
                    \citep{Lv2023} & 2023 & MPC & data-driven neural   predictor & X &  &  & X &  &  &  &  \\
                    \citep{Peng2019PathGuided} & 2019 & Adaptive & NN & X & X &  & X &  &  &  &  \\
                    \citep{Sun2022} & 2022 & Sliding-modes & a three-layer NN & X & X &  &  &  &  &  &  \\
                    \citep{Wu2022} & 2022 & BLF & recurrent neural   networks (RNNs) &  &  &  &  &  & X &  &  \\
                    \citep{Wu2023} & 2023 & BLF & recurrent neural   networks (RNNs) &  &  &  &  &  & X &  &  \\
                    \citep{Zhang2021LVS} & 2021 & Adaptive & NN-observer &  &  &  & X & X &  & X &  \\
                    \citep{Zhou2022} & 2022 & Adaptive & NN & X &  &  &  &  &  &  & \\
        \bottomrule
    \end{tabular}}
\label{fig: Machine Learning}
\end{table*}

Below, we include a summary of the approaches taken by various papers, considering machine learning techniques and how each one addresses different problems. In general, the following techniques were identified and grouped:

\begin{itemize}
    \item Neural networks, which encompass techniques such as Wavelet Neural Network (WNN) or Single Hidden Layer Neural Networks (SHLNN), among others.
    \item Radial Basic Function Neural Networks (RBF-NN), which, although could be grouped within the previous category, have been treated individually due to their widespread use.
    \item Minimal Learning Parameter (MLP)
    \item Reinforcement Learning, being the least utilized.
\end{itemize}

Taking a closer look at the challenges faced by machine learning techniques, it is evident that model uncertainties constitute over 80\%, being the most common objective in the use of machine learning. However, each article takes a different approach. For instance, \citep{Ghommam2020} proposes a two-layer approach, with the first layer handling distributed control and a second layer managing force and torque control. In this case, adaptive control is implemented, where local neural networks are capable of estimating parameters to enhance behavior and performance. This approach is quite similar to that of \citep{He2021}, where a neural network is also proposed to adjust error weights based on velocity measurements, enhancing the performance of Dynamic Surface Control (DSC).

The second most frequently addressed challenge by machine learning techniques is external disturbances, which, in most cases, are grouped and simultaneously considered with model uncertainties, as disturbances can directly affect the latter. However, as mentioned in section \ref{subsubsec:uncertainties}, disturbances can vary depending on their environment and affect fleet behavior differently. Therefore, works such as \citep{Xiong2022}, where a neural network with a filter was implemented to exclusively identify disturbances, or \citep{Ma2019}, where an adaptive neural network is proposed to estimate and pre-compensate for them in control, highlights the importance and contributes to understanding the impact of these disturbances on fleet control.

In comparison to other neural networks, Radial Basic Function Neural Networks have been widely utilized in fleet control, particularly in dynamic control layer, due to their processing time, low susceptibility to overfitting, and ability to identify unknown nonlinear functions. For instance, in \citep{Ning2022}, this technique is employed to estimate model uncertainties and external marine disturbances. However, this technique is rarely used alone and; in many cases, it is combined with other techniques to increase its robustness and improve its performance. For example, in \citep{Zhang2022Bearing}, the MLP method is additionally utilized to effectively reduce computational load by estimating the upper limit of the ideal weight matrix instead of updating the entire matrix.

Another problem closely related to model uncertainties is the determination of the input gain, mathematically defined as the reciprocal of inertial terms related to mass and directly impacting the design of kinetic controllers. Although it can be encompassed as an uncertainty, some papers have addressed it separately. For instance, in \citep{Peng2019PathGuided}, a neural estimator was developed to identify it, while in \citep{Gu2022}, it is calculated using RBF-NN alongside model uncertainties simultaneously. In both cases, this allows for more efficient control design and better performance.

As mentioned previously, most articles utilize machine learning techniques to identify parameters or disturbances in real-time. However, a group of works with a different approach was also identified, where machine learning techniques played a more direct role in fleet control. For instance, in \citep{Liang2020Distributed}, a distributed tracking controller based on WNN is proposed, while in \citep{He2022}, knowledge acquired through RBF-NN is reused to develop decentralized formation control. However, it is noteworthy to mention the proposal of \citep{Zhao2021}, who employs a formation control model based on Deep Reinforcement Learning (DRL), being the only one with this approach in this review.

Another interesting approach is the use of machine learning techniques to compute control actions for actuators. Unlike previously discussed approaches, these methods do not contribute to identifying unknown parameters but rather aim to solve calculation problems while improving execution times. Examples of this approach can be found in \citep{Wu2022} and \citep{Wu2023}, where single-layer recurrent neural networks are employed to solve quadratic optimization and enable real-time implementation. Moreover, some studies utilize multiple machine learning techniques, allowing them to address various problems simultaneously. For instance, in \citep{Chen2022reinforcement}, a neural network is used to identify actuator faults, and reinforcement learning is applied to ensure formation performance without violating speed constraints.

On the other hand, an innovative yet underutilized approach in the analyzed works is the use of machine learning techniques to improve or correct trajectories. In most studies, trajectories or paths are predefined and must adhere to certain constraints to facilitate tracking. However, a different approach is proposed in \citep{chen2019}, which employs a BLS to smooth the trajectories generated by an upper layer. This layer alone is incapable of accounting for the dynamic characteristics of the vessel or external disturbances. Still, through coupling with the BLS, it can identify model uncertainties, external disturbances, and further smooth the trajectories sufficiently to meet the necessary requirements simultaneously.

To conclude, the last approach found was the use of machine learning techniques to estimate not only uncertainties or disturbances but also to reconstruct velocities or positions from measurements of different sensors. This means that machine learning techniques not only serve to support or enhance controller performance but also include them as a fundamental part of control architectures. For example, \citep{Xia2021} proposes a observer based on radial basis function neural networks capable of reconstructing velocities from position and angle measurements. Another similar case is \citep{Zhang2021LVS}, which implemented an observer based on neural networks to reconstruct follower velocities, reducing communication load while also capable of detecting actuator faults and compensating input gains.

\section{Detected Gaps and Future Research} \label{sec:gap_future_research}

We have identified several key areas that require further exploration. Despite discussing these issues in relevant sections, it is crucial to emphasize them here to outline future research directions.

A notable observation is the scarcity of experimental studies within the domain; out of the 125 papers reviewed, merely 17 delve into experimental validations. This disparity underscores a critical need for more empirical investigations to bridge the gap between theoretical advancements and their practical applicability.

Moreover, the literature demonstrates that asynchronous communication among ASVs has received limited attention. Given the dynamic and often unpredictable nature of marine environments, enhancing asynchronous communication capabilities could significantly improve the robustness and adaptability of coordinated ASV operations. Some proposals for future research include the development of more robust and adaptive communication protocols utilizing advanced networking technologies such as mobile networks and satellite communication, which can be implemented in more challenging environments and allow for the assessment of their impact on fleet controller performance.

Among the experimental studies, a conspicuous deficiency is the lack of validation in diverse and challenging environments, such as rivers or areas with strong current perturbations. Such conditions are commonplace in real-world scenarios and pose significant challenges to ASV coordination, necessitating more focused research in these contexts.

The exploration of ASV designs in the reviewed literature appears to be narrowly focused, with minimal consideration given to alternative configurations like tricatamarans. Expanding the scope to include a wider variety of vessel designs could unveil novel control strategies and operational efficiencies, enriching the field with diverse perspectives.

Another aspect that requires attention in future work is the importance and impact of communications between fleet agents. Factors such as latency and link reliability are essential in experimental validation and can have a considerable impact on controller performance. Similarly, another aspect is communication security and information encryption to protect system integrity. In real-world applications, communication between agents becomes an essential factor that must be evaluated and validated experimentally.

The issue of fault detection and actuator blockage resolution in experimental validations is another critical gap. The ability to detect and mitigate failures is paramount for ensuring the safety and reliability of ASV fleets, particularly in complex coordination scenarios. Thus, this area emerges as a fertile ground for future research, aiming to enhance the resilience of ASV operations.

Although machine learning techniques have been used in coordinated control schemes, there is a clear gap in the application of techniques such as deep learning, reinforcement learning, and deep reinforcement learning. These methods enable the proposal of deeper, feedback-based networks that enhance the robustness of control architectures. In addition to being used for parameter estimation or disturbance management, they can also be integrated into coordinated control for decision-making tasks, such as obstacle avoidance or generating speed and orientation references.


Finally, there is a significant deficiency in validations involving a larger number of vehicles. Increasing the number of coordinated vessels in experimental setups could provide valuable data on the complexity and scalability of control algorithms. Additionally, this would allow for the assessment of the impact of a larger fleet on communication and the quality of information shared among agents. Factors such as medium access, communication interference, and the need for increased information traffic should be experimentally validated in larger fleets. This would not only allow for the evaluation of the scalability of communication architectures but also their direct impact on the performance of control algorithms.

In summary, while the reviewed literature lays a solid foundation in the field of ASV coordination, the detected gaps highlight substantial opportunities for future research.

At this link (\url{https://hdl.handle.net/20.500.12412/5698}), an Excel document is provided, compiling all extracted features for each article in one comprehensive table. It is made available not only for consultation but also as a foundation for future expansion and updates.

\section{Conclusions}\label{sec:conclusions}

This paper offers a systematic review of the latest advancements in the coordinated control of multiple ASVs. Our study centers on underactuated surface vessels, delving into coordinated control strategies that are customized to their distinctive features. We delve into various methods, outlining their advantages and disadvantages, applications, and the specific coordinated control tactics applied in each case. Furthermore, we investigate the techniques that integrate machine learning for improved autonomous capabilities. Several gaps have been identified, highlighting areas that are ripe for future exploration.

\section*{CRediT authorship contribution statement}

\textbf{Manuel Gantiva Osorio:} Methodology, Formal analysis, Data Curation, Writing - Original Draft, Investigation \textbf{Carmelina Ierardi:} Conceptualization, Methodology, Formal analysis, Data Curation, Writing - Original Draft \textbf{Isabel Jurado Flores:} Methodology, Data Curation, Writing - Original Draft \textbf{Mario Pereira Martín:} Methodology, Data Curation, Writing - Original Draft \textbf{Pablo Millán Gata:} Conceptualization, Methodology, Data Curation, Writing - Original Draft y Funding acquisition.

\section*{Data availability}

The data resulting from this systematic review has been compiled in this link. (\url{https://hdl.handle.net/20.500.12412/5698}).

\bibliographystyle{cas-model2-names}






\end{document}